\pdfoutput=1

\documentclass[runningheads]{llncs}
\usepackage{graphicx}

\usepackage{tikz}
\usepackage{comment}
\usepackage{amsmath,amssymb} 
\usepackage{color}

\usepackage[accsupp]{axessibility}  

\usepackage{amsfonts}
\usepackage{bm}
\usepackage{booktabs}
\usepackage{enumerate}
\usepackage{epsfig}
\usepackage{makecell}
\usepackage{mathrsfs}
\usepackage{microtype}
\usepackage{multirow}
\usepackage{nicefrac}

\begin{document}
\pagestyle{headings}
\mainmatter

\title{Learning Object Placement via Dual-path Graph Completion} 

\titlerunning{Learning Object Placement}
%
\author{Siyuan Zhou \and
Liu Liu \and
Li Niu* \and Liqing Zhang}
\authorrunning{S. Zhou et al.}
%
\institute{MoE Key Lab of Artificial Intelligence, Shanghai Jiao Tong University, China
\email{\{ssluvble,Shirlley,ustcnewly\}@sjtu.edu.cn}, \email{zhang-lq@cs.sjtu.edu.cn}}

\maketitle

\let\thefootnote\relax\footnotetext{*Corresponding Author}

\begin{abstract}
Object placement aims to place a foreground object over a background image with a suitable location and size. In this work, we treat object placement as a graph completion problem and propose a novel graph completion module (GCM). The background scene is represented by a graph with multiple nodes at different spatial locations with various receptive fields. The foreground object is encoded as a special node that should be inserted at a reasonable place in this graph. We also design a dual-path framework upon the structure of GCM to fully exploit annotated composite images. With extensive experiments on OPA dataset, our method proves to significantly outperform existing methods in generating plausible object placement without loss of diversity. \emph{Codes are avaliable at: https://github.com/bcmi/GracoNet-Object-Placement.}
\end{abstract}

\section{Introduction}\label{sec:intro}

Image composition~\cite{chen2019toward,weng2020misc,niu2021making} aims to produce a realistic composite image based on a background image and a foreground object, which can benefit a wide range of applications of entertainment, virtual reality, and artistic creation. The main concerns of this task include appearance compatibility (\emph{e.g.}, shading, lighting), geometric compatibility (\emph{e.g.}, object size, camera viewpoint), and semantic compatibility (\emph{e.g.}, semantic context) between foreground and background~\cite{chen2019toward,lin2018st}. In this work, we deal with object placement~\cite{lee2018contextaware,tripathi2019learning,zhang2020learning}, which is a sub-task of image composition and aims to generate reasonable locations and sizes to place foreground over background. Object placement can be applied in various conditions. For example, during artistic creation, this technique could provide designers with feedback and suggestions when they are placing objects. Another application is automatic advertising, which aims to help advertisers insert product in the background scene~\cite{zhang2020and}.
Object placement is a challenging problem, partially due to the lack of annotated composite images. Recently, the first object placement assessment (OPA) dataset~\cite{liu2021opa} was released, which contains composite images and their binary rationality labels indicating whether they are reasonable (positive sample) or not (negative sample) in terms of foreground object placement.

\begin{figure}[t]
	\centering
	\includegraphics[width=0.71\linewidth]{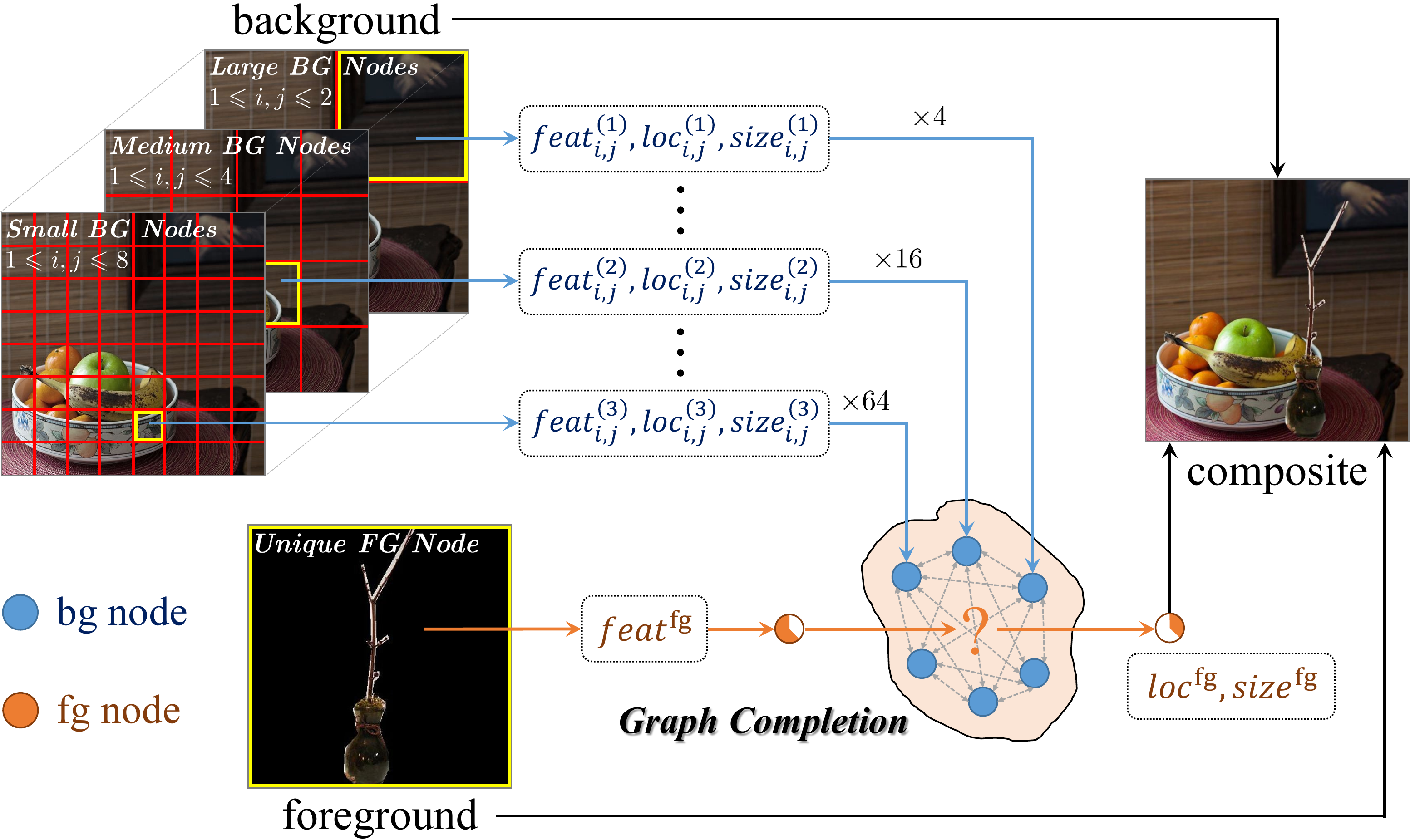}
	\caption{Illustration of our graph completion module (GCM). Background nodes are extracted from different positions with different receptive fields. The unique foreground node lacks location and size information. GCM infers the missing information of the foreground node to complete the graph}
	\label{fig:intro}
\end{figure}

As far as we know, only few works focus on general object placement learning, like TERSE~\cite{tripathi2019learning} and PlaceNet~\cite{zhang2020learning}. Both of them adopt adversarial training to learn the reasonable distribution from real images, and predict a set of parameters to indicate locations and sizes for placing foreground objects during inference. The drawbacks of these methods mainly come from two aspects. Firstly, they did not explicitly consider the relation between the foreground object and the background scene, which is of great importance in object placement. Secondly, they did not fully exploit the annotated composite images. Based on these considerations, we design our method to generate more plausible and diverse object placements.

To better exploit the relationship between the foreground object and the background scene, we treat object placement learning as a graph completion problem, as shown in Figure~\ref{fig:intro}. On the one hand, the background image can be considered as a graph with multiple nodes. Each node pays attention to a local region on the background feature map with a specific spatial position and receptive field. Multiple background nodes work together to form a graph, enabling the model to discover a variety of plausible solutions on the background. On the other hand, the foreground image can be seen as a special node that should be inserted into the graph with suitable location and size. Note that the background nodes have both feature information and location/size information, whereas the foreground node only has feature information. We need to infer the missing location/size information for the foreground node to complete the graph and obtain a reasonable composite layout.

To complete the graph, we propose a novel graph completion module (GCM) with two components: node extraction head (NEH) and placement seeking network (PSN). NEH aims to transform foreground and background into a node graph. We incorporate one foreground NEH to extract a foreground node and another background NEH to extract multiple background nodes from different positions and scales. PSN contains an attention layer and a regression block. The attention layer attends relevant information from different background regions for the foreground object and produces an attended feature vector, which will be transmitted to the regression block to predict transformation parameters for object placement. Note that object placement is a multi-modal problem, that is, the reasonable placement has many possible solutions given a pair of foreground and background. This guides us to incorporate a random vector in the regression block to generate diversified transformation parameters for object placement. 

To take full advantage of annotated composite images, we design a dual-path framework upon the structure of GCM, including an unsupervised path and a supervised path. The whole framework follows an adversarial learning paradigm, which is composed by a GCM (functioning as a generator) and a discriminator. Transformation parameters produced by GCM are applied to predict reasonable object placements, which are then pushed to the discriminator so as to check the plausibility of the generated composite images. The distinction between two paths lies in the provided data. The unsupervised path only utilizes pairs of foreground and background as input, while the supervised path have additional annotated composite images. Recall that GCM contains a random vector in the regression block. In our implementation, two paths choose different types of random vectors. In the unsupervised path, random vectors are sampled from unit Gaussian distribution. In the supervised path, random vectors (also called latent vectors) are encoded from composite images with positive annotation via a VAE~\cite{kingma2014autoencoding} encoder. We expect the generator to reconstruct the ground-truth transformation parameters of each positive composite image from its corresponding latent vector. Under this design, we establish a bijection between the latent vector and the predicted object placement. This can avoid mode collapse~\cite{zhu2017multimodal} and bring multifarious generation results. Since two paths share weights in the generator and the discriminator, we hope that the supervised path could gradually guide the unsupervised path to generate reasonable composite images. During inference, by sampling random vectors in the unsupervised path, we can obtain diverse solutions for object placement. Since the key idea of this work is \textbf{Gra}ph \textbf{co}mpletion, we name our network \textbf{GracoNet}. 

In summary, the main contributions of this paper are: 1) We formulate object placement as a graph completion problem and propose a novel graph completion module (GCM). 2) We design a dual-path framework upon GCM to fully exploit annotated composite images and overcome mode collapse issue. 3) Experiments on OPA dataset demonstrate the superiority of our method in generation plausibility and diversity when compared with existing works.

\section{Related Work}\label{sec:related}

\subsection{Image Composition}\label{subsec:imgcomp}

The main challenges of image composition~\cite{niu2021making,smith1996blue,lalonde2007using,xue2012understanding,lalonde2007photo,chen2009sketch2photo} lie in appearance compatibility, geometric compatibility, and semantic compatibility between foreground and background. Up to now, this task has been explored from a variety of perspectives.
For example,~\cite{zhu2015learning} refined composite images by distinguishing them from natural photographs via a simple CNN model.~\cite{johnson2018image} incorporated scene graphs to explicitly learn relationships between objects and generate images from a computed scene layout.~\cite{chen2019toward} introduced a new GAN architecture to explore geometric and color correction at the same time.~\cite{weng2020misc} pointed out the drawback of cutting-edge methods and addressed it by a spatially-adaptive mechanism.~\cite{wu2019gp,zhang2020deep} explored the image blending field and achieved seamless connection between foreground and background via blending boundary regions.~\cite{liu2020arshadowgan,hong2022shadow} generated realistic shadows \emph{w.r.t} foreground objects over background scenes.~\cite{tsai2017deep,cong2020dovenet,cong2021bargainnet,cong2022high} proposed image harmonization to deal with color and lighting inconsistency in composite images. Additionally, object placement has been studied to realize geometric compatibility, which will be introduced next.

\subsection{Object Placement}\label{subsec:objplace}

Learning object placement has attracted wide attention in recent years. Several early methods~\cite{schuster2010perceiving,georgakis2017synthesizing} attempted to design explicit rules to place foreground objects. The followers went a step further to automatically exploit reasonable placement~\cite{tan2018and,lin2018st,lee2018contextaware,tripathi2019learning,li2019putting,zhang2020learning,zhang2020and,azadi2020compositional}.
For example,~\cite{lin2018st} employed spatial transformer networks to learn geometric corrections that warp composite images for appropriate layouts.~\cite{lee2018contextaware} designed a two-step strategy to find where to place objects and what categories to place.~\cite{li2019putting} used VAE~\cite{kingma2014autoencoding} to predict 3D locations and poses of humans.~\cite{azadi2020compositional} achieved self-consistency in training a composition network by decomposing composite images back into individual objects. Compared with existing works, we offer a new perspective by treating object placement as a graph completion problem. Our dual-path framework could effectively boost generation plausibility and diversity by discovering placement clues from supervisions.

\section{Methodology}\label{sec:method}

Suppose we have a background image $\mathrm{I}^{\mathrm{bg}}\in\mathcal{R}^{3\times H\times W}$ and a foreground image $\mathrm{I}^{\mathrm{fg}}\in\mathcal{R}^{3\times H\times W}$ together with a binary object mask $\mathrm{M}^{\mathrm{fg}}\in\mathcal{R}^{1\times H\times W}$ delineating the foreground object, where $H$ and $W$ represent image height and width. Our objective is to output transformation parameters $\mathbf{t}$ that transform foreground and places it over background to obtain a composite image $\mathrm{I}^{\mathrm{c}}\in\mathcal{R}^{3\times H\times W}$ with a composite foreground mask $\mathrm{M}^{\mathrm{c}}\in\mathcal{R}^{1\times H\times W}$. By using $\mathcal{F}_{\mathbf{t}}$ to represent the transformation function with parameters $\mathbf{t}$, we have \begin{eqnarray}\label{eqn:trans}(\mathrm{I}^{\mathrm{c}},\mathrm{M}^{\mathrm{c}})=\mathcal{F}_{\mathbf{t}}(\mathrm{I}^{\mathrm{bg}}, \mathrm{I}^{\mathrm{fg}}, \mathrm{M}^{\mathrm{fg}}).\end{eqnarray}
The detailed definitions of $\mathbf{t}$ and $\mathcal{F}_{\mathbf{t}}$ are left to supplementary. In the following paragraphs, we will first introduce how GCM works to generate transformation parameters $\mathbf{t}$ in Section~\ref{subsec:gcm}. Then, we will introduce the GCM-based dual-path framework that reasonably makes use of supervised information in Section~\ref{subsec:dualpath}.

\begin{figure*}[t]
	\centering
	\includegraphics[width=\linewidth]{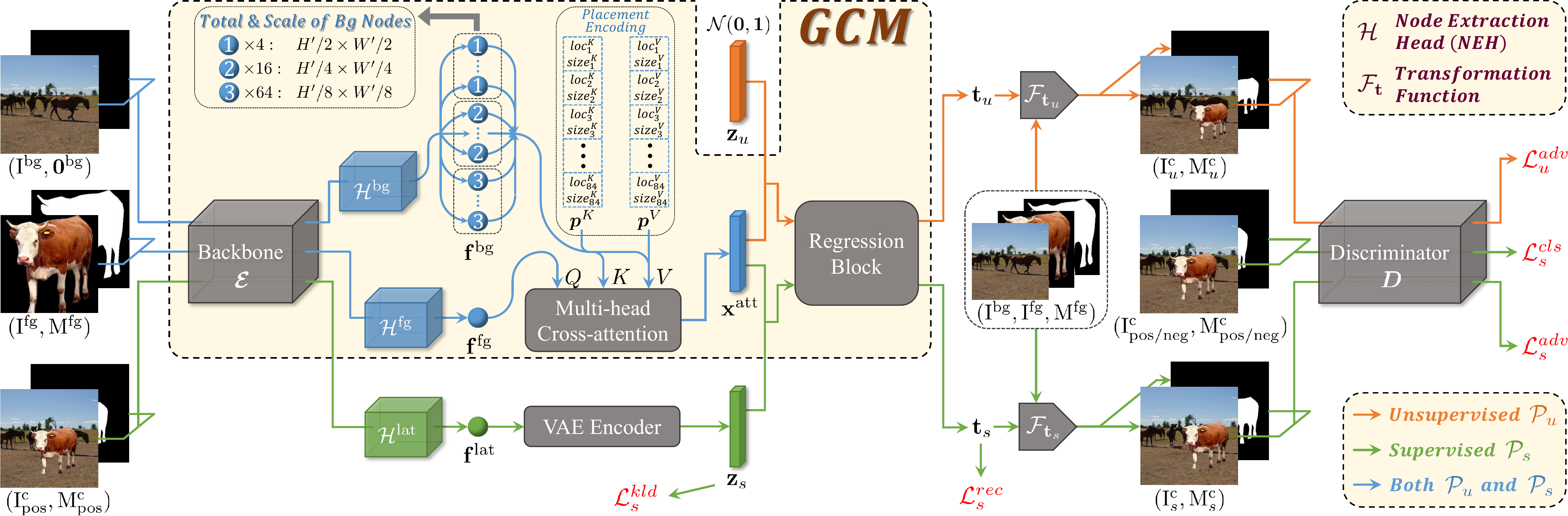}
	\caption{Our GracoNet has an unsupervised path and a supervised path built upon Graph Completion Module (GCM). GCM consists of backbone network $\mathcal{E}$, node extraction head $\mathcal{H}$, and placement seeking network $\mathcal{S}$ (a multi-head cross-attention layer and a regression block).  Loss functions are marked in red. More details are left to Section~\ref{sec:method}}
	\label{fig:net}
\end{figure*}

\subsection{Graph Completion Module (GCM)}\label{subsec:gcm}
The core module in our network is Graph Completion Module (GCM), which takes in a pair of foreground and background as well as a random vector to produce reasonable placement (transformation parameter) for the foreground object.  
GCM consists of three components: 1) backbone network $\mathcal{E}$, 2) node extraction head $\mathcal{H}$, and 3) placement seeking network $\mathcal{S}$. The backbone network extracts general feature maps for the input images. The node extraction head encodes graph nodes from the feature maps. After that, the placement seeking network finds the relationships between the foreground node and the background nodes, and finally outputs transformation parameters $\mathbf{t}$ that reasonably places the foreground node to complete the graph. 

\subsubsection{Backbone Network.}\label{subsubsec:backbone}

Our backbone network $\mathcal{E}$ takes in the concatenation of a three-channel image and a one-channel mask to produce a feature map $\mathrm{F}\in\mathcal{R}^{C'\times H'\times W'}$. We use object masks $\mathrm{M}^{\mathrm{fg}}$ for foreground images $\mathrm{I}^{\mathrm{fg}}$ and apply all-zero masks $\mathbf{0}^{\mathrm{bg}}$ to background images $\mathrm{I}^{\mathrm{bg}}$. Formally, foreground and background features are extracted by $\mathrm{F}^{\mathrm{fg}}=\mathcal{E}(\mathrm{I}^{\mathrm{fg}}, \mathrm{M}^{\mathrm{fg}})$ and $\mathrm{F}^{\mathrm{bg}}=\mathcal{E}(\mathrm{I}^{\mathrm{bg}}, \mathbf{0}^{\mathrm{bg}})$.

\subsubsection{Node Extraction Head (NEH).}\label{subsubsec:neh}

Given a feature map $\mathrm{F}$ as input and a positive integer array $\mathbf{n}=[n_1,n_2,\cdots,n_L]$ as parameter, a node extraction head $\mathcal{H}_{\mathbf{n}}$ consists of $L$ node extraction layers named $\mathcal{G}_1,\mathcal{G}_2,\cdots,\mathcal{G}_L$. The $l$-th layer $\mathcal{G}_l$ evenly divides the feature map $\mathrm{F}$ into $n_l\times n_l$ cells and sequentially encodes them into a stack of $n_l^2$ nodes with dimension $C$, denoted by $\mathbf{f}^{(l)}\in\mathcal{R}^{{n_l^2}\times C}$. Each node accounts for a spatial resolution of $\frac{H'}{n_l}\times \frac{W'}{n_l}$ on the feature map. $\mathcal{H}_\mathbf{n}$ gathers the outputs from all $L$ layers, and produces $\mathbf{f}=[\mathbf{f}^{(1)}, \mathbf{f}^{(2)}, \cdots, \mathbf{f}^{(L)}] \in \mathcal{R}^{N\times C}$ with totally $N=\sum_{l=1}^{L}{n_l^2}$ nodes. Formally, the workflow of $\mathcal{H}_{\mathbf{n}}$ is denoted by $\mathbf{f}=\mathcal{H}(\mathrm{F};\mathbf{n})$. More implementation details of NEH can be found in Section~\ref{subsubsec:detail}.

In GCM, we incorporate a foreground head and a background head, as illustrated in Figure~\ref{fig:net}. The foreground head $\mathcal{H}_{\mathbf{n}=[1]}^{\mathrm{fg}}$ encodes a global foreground node $\mathbf{f}^{\mathrm{fg}}\in\mathcal{R}^{1\times C}$ from the foreground feature map $\mathrm{F}^{\mathrm{fg}}$, \emph{i.e.}, $\mathbf{f}^{\mathrm{fg}}=\mathcal{H}^{\mathrm{fg}}(\mathrm{F}^{\mathrm{fg}};[1])$. Meanwhile, the background head $\mathcal{H}_{\mathbf{n}=[2,4,8]}^{\mathrm{bg}}$ produces $N=84$ local background nodes $\mathbf{f}^{\mathrm{bg}}\in\mathcal{R}^{84\times C}$ from three scales at different locations, \emph{i.e.}, $\mathbf{f}^{\mathrm{bg}}=\mathcal{H}^{\mathrm{bg}}(\mathrm{F}^{\mathrm{bg}};[2,4,8])$. 
On the whole, the unique foreground node and all the 84 background nodes work together to form a node graph, as shown in Figure~\ref{fig:intro}.

\subsubsection{Placement Seeking Network (PSN).}\label{subsubsec:psn}

It is noteworthy that the node graph is now incomplete, because the location/size of the foreground node awaits to be determined. To address this issue, we introduce a placement seeking network $\mathcal{S}$, which consists of a multi-head cross-attention layer and a regression block. 

First, we use a Transformer multi-head attention layer~\cite{vaswani2017attention} to explore the relationship between the unique foreground node $\mathbf{f}^{\mathrm{fg}}\in\mathcal{R}^{1\times C}$ and the 84 local background nodes $\mathbf{f}^{\mathrm{bg}}\in\mathcal{R}^{84\times C}$ by treating $\mathbf{f}^{\mathrm{fg}}$ as query and $\mathbf{f}^{\mathrm{bg}}$ as key/value (\emph{i.e.}, cross-attention). Inspired by Transformer that incorporates position encoding~\cite{gehring2017convolutional,vaswani2017attention,liu2020learning,shaw2018selfattention,dai2019transformerxl,raffel2020exploring} into the attention layer, we introduce placement encoding to encapsulate both location and size information of $\mathbf{f}^{\mathrm{bg}}$. Since different background nodes have distinct positions/scales, they should have different placement encodings.
In our implementation, placement encoding includes a learnable $\bm{p}^K$ (\emph{resp.}, $\bm{p}^V$) for key (\emph{resp.}, value), which is based on but not exactly the same as the encoding form in~\cite{shaw2018selfattention}. In general, we denote the output of the attention layer by $\mathbf{x}^{\mathrm{att}}=Attention(\mathbf{f}^{\mathrm{fg}},\mathbf{f}^{\mathrm{bg}})\in\mathcal{R}^{1\times C}$, and we will introduce the details as follows.

Specifically, we calculate the output $\bm{o}\in\mathcal{R}^{1\times d_{\bm{o}}}$ of each attention head: \begin{eqnarray}\label{eqn:att_out}\bm{o}=\sum_{j=1}^{84}{\alpha_j(\mathbf{f}^{\mathrm{bg}}_{j}W^V+\bm{p}^V_j)},\end{eqnarray} where coefficient $\alpha_j$ represents the edge weight between the foreground node and the $j$-th background node in the graph: \begin{eqnarray}\label{eqn:att_alpha}\alpha_j=\mathrm{Softmax}\bigg(\frac{(\mathbf{f}^{\mathrm{fg}}W^Q)(\mathbf{f}^{\mathrm{bg}}_{j}W^K+\bm{p}^K_j)^\top}{\sqrt{d_{\bm{o}}}}\bigg).\end{eqnarray} In Eqn.(\ref{eqn:att_out}) and Eqn.(\ref{eqn:att_alpha}), $W^Q,W^K,W^V\in\mathcal{R}^{C\times d_{\bm{o}}}$ are linear learnable weights for query, key, and value, respectively. $\bm{p}^K,\bm{p}^V\in\mathcal{R}^{84\times d_{\bm{o}}}$ are learnable placement encodings. Following~\cite{vaswani2017attention}, we incorporate 8 attention heads and set $d_{\bm{o}}$ as $\frac{C}{8}$ in our implementation. $\bm{p}^K$ and $\bm{p}^V$ are not shared among different attention heads. In this way, different attention heads could potentially discover various placement information, resulting in more diversified generation results. The attention output $\bm{o}$ from different heads are concatenated and transformed with another linear layer to obtain the final output $\mathbf{x}^{\mathrm{att}}\in\mathcal{R}^{1\times C}$ of the attention layer.

Second, we apply a regression block to predict transformation parameters from the attention output. In order to generate composite images with diversified reasonable placements, we incorporate a random vector $\mathbf{z}\in\mathcal{R}^{1\times C_z}$ with dimension $C_z$ into the block. Specifically, the regression block takes in the concatenation of $\mathbf{x}^{\mathrm{att}}$ and $\mathbf{z}$ to predict  transformation parameters $\mathbf{t}=Regression(\mathbf{x}^{\mathrm{att}},\mathbf{z})$. By sampling different $\mathbf{z}$ at test time, we can obtain a variety of reasonable $\mathbf{t}$ conditioned on $\mathbf{x}^{\mathrm{att}}$. The detailed implementation for $\mathbf{z}$ is left to Section~\ref{subsec:dualpath}.

\subsection{Dual-path Framework}\label{subsec:dualpath}
Our whole framework is designed as a Generative Adversarial Network (GAN)~\cite{goodfellow2014generative} including a generator $G$ and a discriminator $D$. The generator $G$ is comprised of a graph completion module (GCM) and a transformation function $\mathcal{F}$. As introduced in Section~\ref{subsec:gcm}, GCM works by predicting transformation parameters $\mathbf{t}$ from a tuple of $(\mathrm{I}^{\mathrm{bg}}, \mathrm{I}^{\mathrm{fg}}, \mathrm{M}^{\mathrm{fg}}, \mathbf{z})$. Using $\mathbf{t}$ as parameters for $\mathcal{F}$, we could follow Eqn.(\ref{eqn:trans}) to obtain a generated composite image $\mathrm{I}^{\mathrm{c}}$ with object mask $\mathrm{M}^{\mathrm{c}}$. Formally, the workflow of our generator $G$ is denoted by $(\mathrm{I}^{\mathrm{c}},\mathrm{M}^{\mathrm{c}})=G(\mathrm{I}^{\mathrm{bg}}, \mathrm{I}^{\mathrm{fg}}, \mathrm{M}^{\mathrm{fg}}, \mathbf{z})$. Then, the discriminator $D$ takes the concatenated $(\mathrm{I}^{\mathrm{c}},\mathrm{M}^{\mathrm{c}})$ as input, and predicts the probability of reasonableness for the generated composite image.

To facilitate object placement learning, we adopt a dual-path adversarial training framework containing an unsupervised path $\mathcal{P}_u$ and a supervised path $\mathcal{P}_s$. In $\mathcal{P}_u$, we only have background images $\mathrm{I}^{\mathrm{bg}}$ and foreground images $\mathrm{I}^{\mathrm{fg}}$ with object masks $\mathrm{M}^{\mathrm{fg}}$. In $\mathcal{P}_s$, we are provided with additional annotated composite images/masks $\mathrm{I}^{\mathrm{c}}_{\mathrm{pos}}$/$\mathrm{M}^{\mathrm{c}}_{\mathrm{pos}}$ (\emph{resp.}, $\mathrm{I}^{\mathrm{c}}_{\mathrm{neg}}$/$\mathrm{M}^{\mathrm{c}}_{\mathrm{neg}}$) with positive (\emph{resp.}, negative) annotation in terms of object placement, as well as their corresponding original $\mathrm{I}^{\mathrm{bg}}$/$\mathrm{I}^{\mathrm{fg}}$/$\mathrm{M}^{\mathrm{fg}}$ that constitute them. Due to the difference of provided data, $\mathcal{P}_u$ and $\mathcal{P}_s$ adopt distinct implementations for the random vector $\mathbf{z}$. To distinguish between representations in two paths, we use notation $\mathbf{z}_u$ in $\mathcal{P}_u$ and $\mathbf{z}_s$ in $\mathcal{P}_s$, respectively. The generated composite outputs in two paths are represented by $\mathrm{I}^{\mathrm{c}}_u$/$\mathrm{M}^{\mathrm{c}}_u$ and $\mathrm{I}^{\mathrm{c}}_s$/$\mathrm{M}^{\mathrm{c}}_s$ with different subscripts correspondingly. 

\subsubsection{Unsupervised Path ($\mathcal{P}_u$).}\label{subsubsec:path_u}

In unsupervised path $\mathcal{P}_u$, random vectors $\mathbf{z}_u$ are sampled from unit Gaussian distribution, \emph{i.e.}, $\mathbf{z}_u\sim\mathcal{N}(\mathbf{0},\mathbf{1})$. Correspondingly, the generator outputs $(\mathrm{I}^{\mathrm{c}}_u,\mathrm{M}^{\mathrm{c}}_u)=G(\mathrm{I}^{\mathrm{bg}}, \mathrm{I}^{\mathrm{fg}}, \mathrm{M}^{\mathrm{fg}}, \mathbf{z}_u)$. We employ an adversarial loss $\mathcal{L}^{adv}_{u}(G,D)$ to push the generated composite image to be undistinguishable from positive composite images. 

\begin{eqnarray}\label{eqn:loss_gan_u}\mathcal{L}^{adv}_{u}=\mathbb{E}_{\mathbf{z}_u\sim \mathcal{N}(\mathbf{0},\mathbf{1})}[\log(1-D(G(\mathrm{I}^{\mathrm{bg}}, \mathrm{I}^{\mathrm{fg}}, \mathrm{M}^{\mathrm{fg}}, \mathbf{z}_u)))].\end{eqnarray}

\subsubsection{Supervised Path ($\mathcal{P}_s$).}\label{subsubsec:path_s}

In supervised path $\mathcal{P}_s$, random vectors (also called latent vectors) $\mathbf{z}_s$ are designed to be sampled from global features of positive composite images via an encoder network as in VAE~\cite{kingma2014autoencoding}. Given a labeled positive composite image $\mathrm{I}^{\mathrm{c}}_{\mathrm{pos}}$ with object mask $\mathrm{M}^{\mathrm{c}}_{\mathrm{pos}}$, we first calculate its ground-truth transformation parameters $\mathbf{t}_{\mathrm{gt}}$ (see details in supplementary), which obeys $(\mathrm{I}^{\mathrm{c}}_{\mathrm{pos}},\mathrm{M}^{\mathrm{c}}_{\mathrm{pos}})\equiv\mathcal{F}_{\mathbf{t}_{\mathrm{gt}}}(\mathrm{I}^{\mathrm{bg}}, \mathrm{I}^{\mathrm{fg}}, \mathrm{M}^{\mathrm{fg}})$. Then, our idea is to employ the latent vector $\mathbf{z}_s$ to help reconstruct $\mathbf{t}_{\mathrm{gt}}$ because $\mathbf{z}_s$ contains potential information of positive composite images. In this way, we establish a bijection between $\mathbf{z}_s$ and $(\mathrm{I}^{\mathrm{c}}_{\mathrm{pos}},\mathrm{M}^{\mathrm{c}}_{\mathrm{pos}})$. 

Specifically, we first use the backbone network $\mathcal{E}$ to extract a feature map  $\mathrm{F}^{\mathrm{c}}=\mathcal{E}(\mathrm{I}^{\mathrm{c}}_{\mathrm{pos}}, \mathrm{M}^{\mathrm{c}}_{\mathrm{pos}})$. Then we add a latent head $\mathcal{H}^{\mathrm{lat}}_{\mathbf{n}=[1]}$ to encode a global latent node $\mathbf{f}^{\mathrm{lat}}=\mathcal{H}^{\mathrm{lat}}(\mathrm{F}^{\mathrm{c}};[1])$. After that, we employ the encoder network in VAE to sample a latent vector $\mathbf{z}_s$ from $\mathbf{f}^{\mathrm{lat}}$. Following VAE, we adopt a KL divergence loss $\label{loss_kld}\mathcal{L}^{kld}_s=\mathcal{D}_{\mathrm{KL}}(\mathcal{N}(\bm{\mu}_{\mathbf{z}_s}, \bm{\sigma}_{\mathbf{z}_s}^2)\parallel\mathcal{N}(0,1))$ that forces the distribution of $\mathbf{z}_s$ to be close to $\mathbf{z}_u$. With $\mathbf{z}_s$, GCM predicts transformation parameters $\mathbf{t}_s$ from a tuple of $(\mathrm{I}^{\mathrm{bg}}, \mathrm{I}^{\mathrm{fg}}, \mathrm{M}^{\mathrm{fg}}, \mathbf{z}_s)$. We utilize a reconstruction loss $\mathcal{L}^{rec}_s$ to force $\mathbf{t}_s$ to approach the ground-truth $\mathbf{t}_{\mathrm{gt}}$, which is defined as a weighted MSE between $\mathbf{t}_s$ and $\mathbf{t}_{\mathrm{gt}}$ (see details in supplementary). Then, transformation function $\mathcal{F}$ with parameters $\mathbf{t}_s$ produces $(\mathrm{I}^{\mathrm{c}}_{s},\mathrm{M}^{\mathrm{c}}_{s})$, which is finally delivered to the discriminator $D$. Similar to $\mathcal{P}_u$, we also adopt an adversarial loss in $\mathcal{P}_s$: \begin{eqnarray}\label{eqn:loss_gan_s}\mathcal{L}^{adv}_{s}=\mathbb{E}_{\mathbf{z}_s\sim \mathcal{N}(\bm{\mu}_{\mathbf{z}_s}, \bm{\sigma}_{\mathbf{z}_s}^2)}[\log(1-D(G(\mathrm{I}^{\mathrm{bg}}, \mathrm{I}^{\mathrm{fg}}, \mathrm{M}^{\mathrm{fg}}, \mathbf{z}_s)))].\end{eqnarray} Additionally, we leverage both positive and negative composite images to update the discriminator by maximizing the negative form of binary cross-entropy loss: \begin{eqnarray}\label{eqn:loss_cls}\begin{aligned}\mathcal{L}^{cls}_s=\log{D(\mathrm{I}^{\mathrm{c}}_{\mathrm{pos}},\mathrm{M}^{\mathrm{c}}_{\mathrm{pos}})}+\log{(1-D(\mathrm{I}^{\mathrm{c}}_{\mathrm{neg}},\mathrm{M}^{\mathrm{c}}_{\mathrm{neg}}))}.\end{aligned}\end{eqnarray} In summary, the loss function for $\mathcal{P}_s$ is defined by \begin{eqnarray}\label{eqn:loss_s}\mathcal{L}_s(G,D)=\mathcal{L}^{kld}_s(G)+\lambda\mathcal{L}^{rec}_s(G)+\mathcal{L}^{adv}_s(G,D)+\mathcal{L}^{cls}_s(D),\end{eqnarray} where the hyper-parameter $\lambda$ is set as 50. Note that $\mathcal{L}^{adv}_s(G,D)$, $\mathcal{L}^{kld}_s(G)$, and $\mathcal{L}^{rec}_s(G)$ only handle positive composite images. 

By using $\theta_G$ and $\theta_D$ to represent learnable weights in $G$ and $D$, our optimization objective in the whole framework is \begin{eqnarray}\label{eqn:opt_s}\min\limits_{\theta_G}\max\limits_{\theta_D}\quad\mathcal{L}^{adv}_u(G,D)+\mathcal{L}_s(G,D).\end{eqnarray} Note that two paths share weights in both $G$ and $D$. Under this design, the supervised path could gradually guide the unsupervised path to generate composite images with reasonable object placement. During inference, we only use the unsupervised path to generate composite images by sampling $\mathbf{z}_u$ from $\mathcal{N}(\mathbf{0},\mathbf{1})$.

\section{Experiment}\label{sec:exp}

\subsection{Experimental Setting}\label{subsec:setting}

\subsubsection{Dataset and Evaluation Metrics.}\label{subsubsec:dataset}

We perform experiments on OPA dataset \cite{liu2021opa}, which provides binary rationality labels for composite images. The dataset includes 62074 (21376 positive / 40698 negative) composite images for training and 11396 (3588 positive / 7808 negative) composite images for testing. These annotated composite images in the dataset contain 1389 different background scenes and 4137 different foreground objects from 47 categories.

Since our objective is to generate composite images with reasonable object placements, we train our model on OPA \emph{train} set and evaluate it on the 3588 positive samples of OPA \emph{test} set. During inference, our model takes the foreground/background of each positive test sample as input, and generates 10 composite images by randomly sampling 10 different $\mathbf{z}_u$ in $\mathcal{P}_u$.

We adopt user study, accuracy, and FID~\cite{heusel2017gans} to evaluate generation plausibility and LPIPS~\cite{zhang2018unreasonable} for generation diversity. User study will be introduced in Section~\ref{subsec:compare}. We extend SimOPA~\cite{liu2021opa} as a binary classifier to distinguish between reasonable and unreasonable object placements. We define accuracy as the proportion of generated composite images that are classified as positive by the binary classifier. FID is calculated between the composite images generated by our method and the positive composite images in the $\emph{test}$ set. We compute LPIPS for all pairs of composite images among 10 generation results for each sample, and adopt the averaged LPIPS among all samples. 

\begin{figure*}[t]
  \centering
  \includegraphics[width=\linewidth]{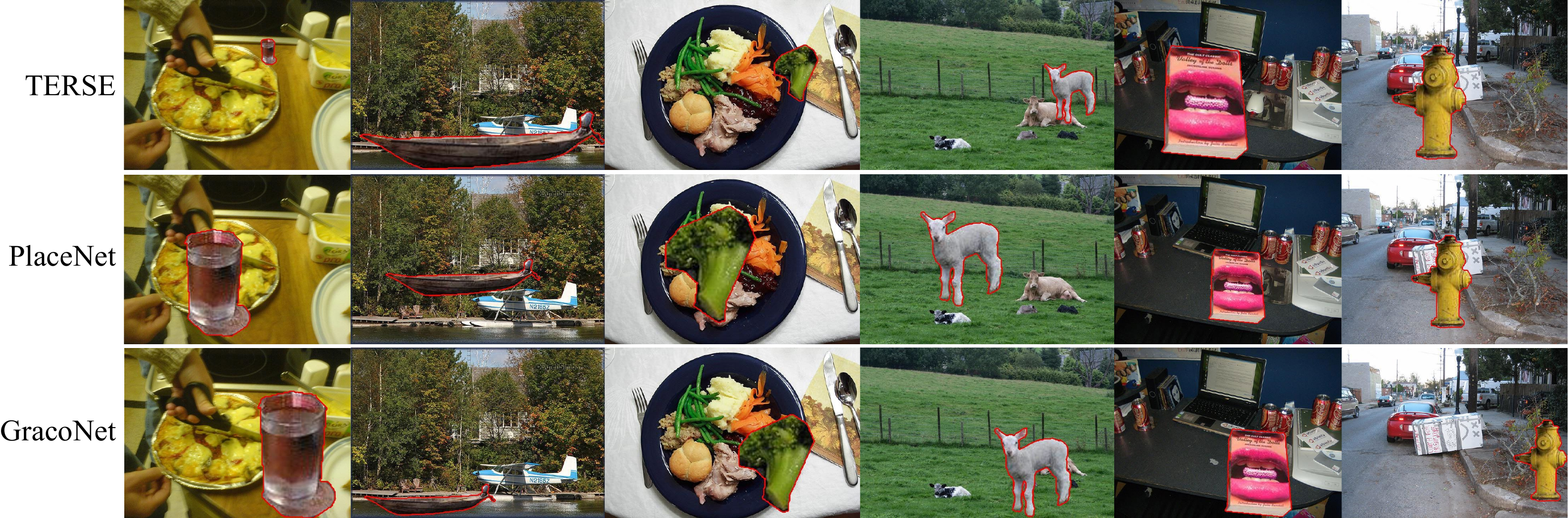}
  \caption{Visualization of object placement results. Foreground is outlined in red}
  \label{fig:fgdiff_bgdiff}
\end{figure*}

\subsubsection{Implementation Details.}\label{subsubsec:detail}

Our backbone network is the beginning 34 layers (including and before the fourth $\mathrm{MaxPool}$ layer) of VGG16~\cite{simonyan2015deep} with batch normalization, except that the first $\mathrm{Conv}$ layer has four input channels. In NEH, each node extraction layer $\mathcal{G}_l$ with parameter $n_l$ contains three groups of ($3\times 3$ $\mathrm{Conv}$, $\mathrm{BN}$, $\mathrm{ReLU}$), followed by a $n_l\times n_l$ $\mathrm{AdaptiveAvgPool}$. The regression block contains three fully connected layers ($\mathrm{Fc1024}$, $\mathrm{Fc1024}$, $\mathrm{Fc3}$), followed by an activation function $(\mathrm{tanh}(\cdot)+1)/2$ that normalizes transformation parameters to range $(0,1)$. We adopt the discriminator architecture in~\cite{wang2018high}. All images are resized to $256\times 256$ and normalized before being fed into the network (see supplementary for more details). The VGG16 backbone is pretrained on ImageNet~\cite{deng2009imagenet}. Our model is trained with batch size 32 for 11 epochs on a single RTX 3090 GPU. We adopt Adam optimizer with $\beta_1=0.5$ and $\beta_2=0.999$ for optimization. The learning rate is initialized as $2 \times 10^{-5}$ for backbone and discriminator, and $2 \times 10^{-4}$ for the remaining parts. For hyper-parameters, we set dimension $C$ as 512 for nodes, and $C_z$ as 1024 for random vectors, which will be carefully analyzed in Section~\ref{sec:hyper}.

\subsection{Comparison with Existing Methods}\label{subsec:compare}

We compare our GracoNet with two baselines: TERSE~\cite{tripathi2019learning} and PlaceNet~\cite{zhang2020learning}. Both of them are re-implemented on OPA dataset~\cite{liu2021opa}. For the first baseline, we retain the synthesizer network and the discriminator of TERSE, and remove the target network. This is because the synthesizer is enough to generate composite images that we need, and we do not need to use the target network for down-stream tasks. For the second baseline, we directly use the complete PlaceNet structure to predict object placement without further adjustment. Since TERSE and PlaceNet had been proposed before OPA dataset was released, these object placement methods did not include negative samples in their method. When we conduct experiments on OPA dataset, unless otherwise stated, we fairly use positive samples and negative samples together for both the baselines and our method. Specifically, we use negative samples in baselines by introducing a binary classification loss following Eqn.(\ref{eqn:loss_cls}) to train the discriminator network.

\begin{table}[t]
  \centering
  \caption{Quantitative object placement results for different methods on OPA dataset}
  \label{table:sota}
  \begin{tabular}{p{25mm}<{\centering}|p{20mm}<{\centering}p{13mm}<{\centering}p{13mm}<{\centering}|p{20mm}<{\centering}}
	\hline
	\multirow{2}*{Method} & \multicolumn{3}{c|}{\emph{Plausibility}} & \multicolumn{1}{c}{\emph{Diversity}} \\
	& user study$\uparrow$ & acc.$\uparrow$ & FID$\downarrow$ & LPIPS$\uparrow$ \\
	\hline
	TERSE~\cite{tripathi2019learning} & 0.214 & 0.679 & 46.94 & 0 \\
	PlaceNet~\cite{zhang2020learning} & 0.249 & 0.683 & 36.69 & 0.160 \\
	GracoNet & 0.537 & 0.847 & 27.75 & 0.206 \\
	\hline
  \end{tabular}
\end{table}

Table~\ref{table:sota} shows the quantitative object placement results for baselines and our proposed method.
Among different evaluation metrics, user study, accuracy, and LPIPS are the most important ones. User study is conducted with 20 voluntary participants by comparing the composite images generated by TERSE, PlaceNet, and our method. For each sample, every participant chooses the method producing the most reasonable composite image. Then, each method is scored by the proportion of participants who choose it. The final score of each method is defined by the averaged score over all samples.

By comparison, our method significantly outperforms TERSE/PlaceNet in generation plausibility (0.537 \emph{v.s.} 0.214/0.249 for user study, 0.847 \emph{v.s.} 0.679/0.683 for accuracy, and 27.75 \emph{v.s.} 46.94/36.69 for FID). Also, our method achieves better LPIPS in generation diversity than PlaceNet. Note that TERSE does not incorporate randomness, so its generation diversity is zero. Generally, our method performs satisfactorily and balances plausibility and diversity well.

Figure~\ref{fig:fgdiff_bgdiff} visualizes some object placement results for different methods. As illustrated, our method works better in predicting foreground locations and sizes by comprehensively analyzing different background regions, verifying the effectiveness of our designed GCM. In the supplementary, we show more visualizations of generation plausibility and diversity from three aspects: 1) combination of an identical background scene and different foreground objects, 2) combination of an identical foreground object and different background scenes, and 3) sampling different random vectors for the same pair of foreground and background. 

\subsection{Ablation Studies}\label{sec:ab}

\subsubsection{Different Choices of Background Head $\mathcal{H}^{\mathrm{bg}}$.}\label{subsec:ab_neh}

Table~\ref{table:ab_neh} displays four choices of parameter $\mathbf{n}$ in the background head $\mathcal{H}^{\mathrm{bg}}$. Since the input size of images is $256\times 256$ and the backbone $\mathcal{E}$ contains four pooling layers, feature maps $\mathrm{F}$ are $16\times 16$ in size.  As default, we choose $\mathbf{n}=[2,4,8]$ to represent large-scale, medium-scale, and small-scale background nodes, which pay attention to different regions on the feature map. In detail, each of the 4 large-scale (\emph{resp.}, 16 medium-scale, 64 small-scale) background nodes focuses on a local receptive field of $8\times 8$ (\emph{resp.}, $4\times 4$, $2\times 2$) region on $\mathrm{F}$. In Table~\ref{table:ab_neh}, we show more choices of parameter $\mathbf{n}$ in $\mathcal{H}^{\mathrm{bg}}$. When $\mathbf{n}=[2]$ or $\mathbf{n}=[2,4]$, the model misses small-scale information, so the placement process lacks details in some local regions. When $\mathbf{n}=[2,4,8,16]$, the model tends to learn pixel-wise knowledge, which is redundant in object placement learning and adversely affects network optimization. Overall, our choice $\mathbf{n}=[2,4,8]$ achieves a good balance.

\begin{table}[t]
  \centering
  \caption{Ablation study on background head $\mathcal{H}^{\mathrm{bg}}$}
  \label{table:ab_neh}
  \begin{tabular}{p{38mm}<{\centering}|p{13mm}<{\centering}p{13mm}<{\centering}|p{20mm}<{\centering}}
	\hline
	\multirow{2}*{$\mathcal{H}^{\mathrm{bg}}$ with parameter $\mathbf{n}$} & \multicolumn{2}{c|}{\emph{Plausibility}} & \multicolumn{1}{c}{\emph{Diversity}} \\
	& acc.$\uparrow$ & FID$\downarrow$ & LPIPS$\uparrow$ \\
	\hline
	$\mathbf{n}=[2]$ & 0.807 & 37.44 & 0.136 \\
	$\mathbf{n}=[2,4]$ & 0.821 & 34.23 & 0.146 \\
	$\mathbf{n}=[2,4,8,16]$ & 0.837 & 25.91 & 0.154 \\
	\hline
	$\mathbf{n}=[2,4,8]$ & 0.847 & 27.75 & 0.206 \\
	\hline
  \end{tabular}
\end{table}

\begin{table}[t]
  \centering
  \caption{Ablation study on different types of learnable placement encodings}
  \label{table:ab_enc}
  \begin{tabular}{p{8mm}<{\centering}p{8mm}<{\centering}p{20mm}<{\centering}|p{13mm}<{\centering}p{13mm}<{\centering}|p{20mm}<{\centering}}
	\hline
	\multirow{2}*{$\bm{p}^K$} & \multirow{2}*{$\bm{p}^V$} & shared across & \multicolumn{2}{c|}{\emph{Plausibility}} & \multicolumn{1}{c}{\emph{Diversity}} \\
	& & heads & acc.$\uparrow$ & FID$\downarrow$ & LPIPS$\uparrow$ \\
	\hline
	& & - & 0.793 & 28.22 & 0.120 \\
	$\checkmark$ & & $\checkmark$ & 0.809 & 28.02 & 0.133 \\
	& $\checkmark$ & $\checkmark$ & 0.844 & 25.14 & 0.127 \\
	$\checkmark$ & $\checkmark$ & $\checkmark$ & 0.851 & 29.09 & 0.170 \\
	$\checkmark$ & & & 0.836 & 26.85 & 0.156 \\
	& $\checkmark$ & & 0.839 & 26.22 & 0.154 \\
	\hline
	$\checkmark$ & $\checkmark$ & & 0.847 & 27.75 & 0.206 \\
	\hline
  \end{tabular}
\end{table}

\begin{table}[t]
  \centering
  \caption{Ablation study on functionality of GCM}
  \label{table:ab_gcm}
  \begin{tabular}{p{25mm}<{\centering}p{20mm}<{\centering}|p{13mm}<{\centering}p{13mm}<{\centering}|p{20mm}<{\centering}}
	\hline
	Fg/Bg Feature & Placement & \multicolumn{2}{c|}{\emph{Plausibility}} & \multicolumn{1}{c}{\emph{Diversity}} \\
	Extractor & Finder & acc.$\uparrow$ & FID$\downarrow$ & LPIPS$\uparrow$ \\
	\hline
	global vector & concat+fc & 0.793 & 33.97 & 0.082 \\
	NEH & concat+fc & 0.828 & 31.36 & 0.144 \\
	\hline
	NEH & PSN & 0.847 & 27.75 & 0.206 \\
	\hline
  \end{tabular}
\end{table}

\begin{table}[t]
  \centering
  \caption{Ablation study on loss functions}
  \label{table:ab_loss}
  \begin{tabular}{p{20mm}<{\centering}|p{13mm}<{\centering}p{13mm}<{\centering}|p{20mm}<{\centering}}
	\hline
	\multirow{2}*{Method} & \multicolumn{2}{c|}{\emph{Plausibility}} & \multicolumn{1}{c}{\emph{Diversity}} \\
	& acc.$\uparrow$ & FID$\downarrow$ & LPIPS$\uparrow$ \\
	\hline
	w/o $\mathcal{L}^{adv}_u$ & 0.790 & 32.64 & 0.038 \\
	w/o $\mathcal{L}^{adv}_s$ & 0.800 & 34.76 & 0.057 \\
	w/o $\mathcal{L}^{cls}_s$ & 0.734 & 34.62 & 0.033 \\
	w/o $\mathcal{L}^{kld}_s$ & 0.844 & 29.26 & 0.199 \\
	w/o $\mathcal{L}^{rec}_s$ & 0.767 & 25.53 & 0.131 \\
	\hline
	Ours & 0.847 & 27.75 & 0.206 \\
	\hline
  \end{tabular}
\end{table}

\subsubsection{Different Types of Learnable Placement Encoding.}\label{subsec:ab_enc}

As discussed in Section~\ref{subsubsec:psn}, we adopt placement encoding $\bm{p}^K$ and $\bm{p}^V$ in the attention layer. In our implementation, $\bm{p}^K$ and $\bm{p}^V$ are both learnable and not shared across different attention heads. In this paragraph, we discuss more variants of placement encoding, as shown in Table~\ref{table:ab_enc}. Without placement encoding, the generation plausibility and diversity witness a considerable decrease, because location and size information of different background nodes are missing under this condition. With a single $\bm{p}^K$ or $\bm{p}^V$, the model performs a little better in plausibility, but the diversity is still unsatisfactory. Using $\bm{p}^K$ and $\bm{p}^V$ together works best. If placement encodings are shared across attention heads, different attention heads could not learn diversified attention regions for object placement, resulting in comparably low generation diversity. Therefore, we choose to use $\bm{p}^K$ and $\bm{p}^V$ together, and make them independent across different attention heads. Comprehensively, our choice considers both plausibility and diversity to achieve the best results.

\subsubsection{Functionality of GCM.}\label{subsec:ab_gcm}

As introduced in Section~\ref{subsec:gcm}, our graph completion module (GCM) consists of two important components: node extraction head (NEH) and placement seeking network (PSN). Table~\ref{table:ab_gcm} provides an ablation study on GCM by replacing NEH or PSN with naive network structures. In Table~\ref{table:ab_gcm}, the first experiment uses split branches to extract global features for foreground and background respectively. Then the two global features are concatenated and regressed with several fc layers to obtain the transformation parameters for object placement. Compared with the first experiment, the second one uses NEH to extract features by considering different background locations and sizes. The background nodes are then averaged and concatenated with the foreground node to predict transformation parameters. The last experiment is our method that combines NEH for feature extraction and PSN for placement finding. Comparing the results of the first two experiments, we find that using NEH to explicitly encode different background locations and sizes is beneficial for object placement. By comparing the results of the last two experiments, PSN works better in discovering relationships between foreground and background, and successfully establishes a reasonable connection between foreground node and background nodes in the node graph. In summary, both NEH and PSN are crucial in our method. They work together to ensure the functionality of our proposed GCM.

\subsubsection{Utility of Different Loss Functions.}\label{subsec:ab_loss}

Table~\ref{table:ab_loss} gives an ablation study on different loss functions. Except for $\mathcal{L}^{kld}_s$, deleting any loss makes the performance drop sharply on plausibility or diversity, especially for $\mathcal{L}^{cls}_s$ and $\mathcal{L}^{rec}_s$. Although our model still performs passably without $\mathcal{L}^{kld}_s$, adding this loss can bring some improvement. Generally, every loss makes up an important part, and they work together to guarantee the effectiveness of our method.

\begin{table}[t]
\begin{minipage}[t]{0.495\textwidth}
  \centering
  \caption{Choices of hyper-parameter $C_z$}
  \label{table:hyper_noise}
  \begin{tabular}{p{10mm}<{\centering}|p{12mm}<{\centering}p{12mm}<{\centering}|p{16mm}<{\centering}}
	\hline
	\multirow{2}*{$C_z$} & \multicolumn{2}{c|}{\emph{Plausibility}} & \multicolumn{1}{c}{\emph{Diversity}} \\
	& acc.$\uparrow$ & FID$\downarrow$ & LPIPS$\uparrow$ \\
	\hline
	256 & 0.807 & 36.04 & 0.151 \\
	512 & 0.838 & 35.53 & 0.175 \\
	2048 & 0.827 & 28.10 & 0.208 \\
	4096 & 0.792 & 27.39 & 0.219 \\
	\hline
	1024 & 0.847 & 27.75 & 0.206 \\
	\hline
  \end{tabular}
\end{minipage}
\begin{minipage}[t]{0.495\textwidth}
  \centering
  \caption{Choices of hyper-parameter $\lambda$}
  \label{table:hyper_rec}
  \begin{tabular}{p{10mm}<{\centering}|p{12mm}<{\centering}p{12mm}<{\centering}|p{16mm}<{\centering}}
	\hline
	\multirow{2}*{$\lambda$} & \multicolumn{2}{c|}{\emph{Plausibility}} & \multicolumn{1}{c}{\emph{Diversity}} \\
	& acc.$\uparrow$ & FID$\downarrow$ & LPIPS$\uparrow$ \\
	\hline
	1 & 0.820 & 23.04 & 0.183 \\
	10 & 0.823 & 29.93 & 0.190 \\
	25 & 0.847 & 28.57 & 0.193 \\
	100 & 0.821 & 24.62 & 0.174 \\
	\hline
	50 & 0.847 & 27.75 & 0.206 \\
	\hline
  \end{tabular}
\end{minipage}
\end{table}

\subsection{Hyper-parameter Analyses}\label{sec:hyper}

\subsubsection{Dimension $C_z$ of Random and Latent Vectors.}\label{subsec:hyper_noise}

As introduced in Section~\ref{subsubsec:detail}, we set the dimension of nodes as $C=512$, and set the dimension of random vector $\mathbf{z}_u$ and latent vector $\mathbf{z}_s$ as $C_z=1024$. The value of $C$ is chosen by considering both hardware resource occupation and model performance. Table~\ref{table:hyper_noise} analyzes different choices of $C_z$ in the range of $[256,4096]$. When $C_z$ increases, accuracy first increases, meets a peak at $1024$, and then decreases. Meanwhile, LPIPS increases and FID decreases generally. For a balanced consideration between plausibility and diversity, we choose $C_z=1024$ in our implementation.

\subsubsection{Coefficient $\lambda$ of Reconstruction Loss.}\label{subsec:hyper_rec}

As discussed in Section~\ref{subsec:ab_loss}, reconstruction loss $\mathcal{L}^{rec}_s$ plays an important role in the supervised path. It helps our model reconstruct ground-truth transformation parameters from positive composite images, enabling the supervised path to guide the unsupervised path through the unified training of two paths. In Eqn.(\ref{eqn:loss_s}), reconstruction loss $\mathcal{L}^{rec}_s$ has a hyper-parameter $\lambda$ indicating the coefficient/weight of this loss during optimization. Table~\ref{table:hyper_rec} displays different choices of  $\lambda$ in the range of $[1,100]$. As can be analyzed, our model achieves comparably good plausibility and diversity when $\lambda=50$, which becomes our default choice in all experiments.

\begin{figure*}[t]
  \centering
  \includegraphics[width=\linewidth]{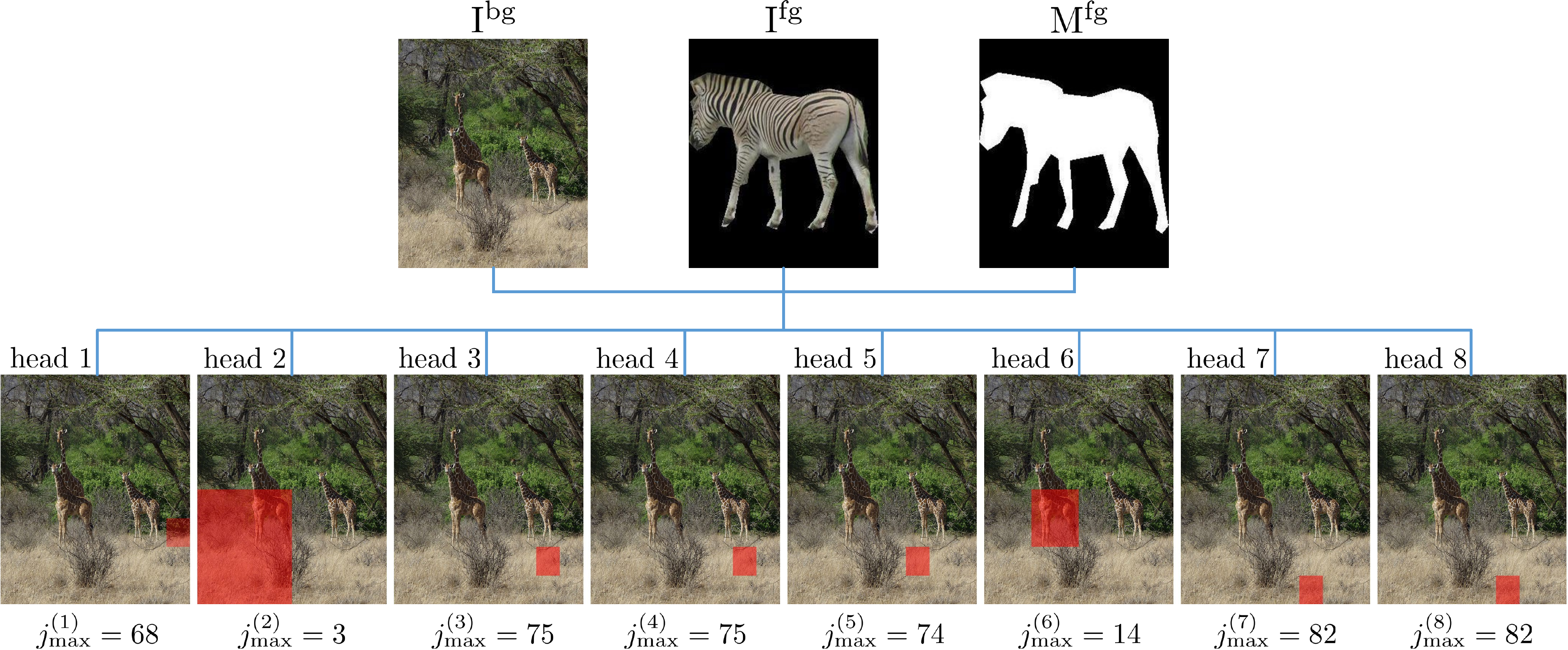}
  \caption{Visualization of background regions (marked in red) that the multi-head attention layer pays the most attention to in each head. $j^{(k)}_{\mathrm{max}}$ is the index of background node with the maximum attention coefficient in the $k$-th head. See details in Section~\ref{subsec:vis_att}}
  \label{fig:att}
\end{figure*}

\subsection{Visualization of Multi-head Attention}\label{subsec:vis_att}

In Section~\ref{subsubsec:psn}, we introduce multi-head attention with placement encoding to discover the relationship between the unique foreground node and totally 84 background nodes. Eqn.(\ref{eqn:att_alpha}) defines attention coefficient $\alpha_j$ ($1\leqslant j\leqslant 84$) in each head to represent the edge weight between the foreground node and the $j$-th background node in the node graph. Specifically, we use $j_\mathrm{max}$ to denote the index with the maximum edge weight: \begin{eqnarray}\label{eqn_jmax}j_\mathrm{max}=\arg\max_{j}{\alpha_j}\quad\quad j=1,2,\cdots,84.\end{eqnarray} Since we have 8 attention heads, we use $j^{(k)}_\mathrm{max}$ with $1\leqslant k\leqslant 8$ for differentiated representations in distinct heads. In the $k$-th attention head, $j^{(k)}_\mathrm{max}$ corresponds to a local background region with a specific location and size encoded by the $j_{\mathrm{max}}$-th background node. According to the definition of background nodes in Section~\ref{subsubsec:neh}, $j^{(k)}_\mathrm{max}$ corresponds to a local background region with scale $\frac{H}{2}\times\frac{H}{2}$ (\emph{resp.}, $\frac{H}{4}\times\frac{H}{4}$, $\frac{H}{8}\times\frac{H}{8}$) when $1\leqslant j^{(k)}_\mathrm{max}\leqslant 4$ (\emph{resp.}, $5\leqslant j^{(k)}_\mathrm{max}\leqslant 20$,\, $21\leqslant j^{(k)}_\mathrm{max}\leqslant 84$). In Figure~\ref{fig:att}, we visualize the local background region attended by $j^{(k)}_\mathrm{max}$ for each attention head, which represents for the region that the $k$-th head pays the most attention to on the background scene. As illustrated, the multi-head attention layer successfully discovers diversified locations and sizes of potential background regions for object placement by learning the relationship between the unique foreground node and multiple background nodes. We could also conclude from Figure~\ref{fig:att} that the most attended regions are reasonable enough for placing the zebra because only near-earth locations are activated, which accords with common sense.

\section{Conclusion}\label{sec:conclusion}

In this work, we have proposed a novel graph completion module (GCM) for the object placement task to explicitly explore the relationship between the foreground object and the background image. We have also designed a dual-path framework upon the GCM structure, in which the supervised path provides additional cues for the unsupervised path so as to significantly enhance the performance. Extensive experiments on OPA dataset have demonstrated the effectiveness of our proposed method.

\subsubsection{Acknowledgements}\label{subsubsec:acknowledgements}

The work is supported by Shanghai Municipal Science and Technology Key Project (Grant No. 20511100300), Shanghai Municipal Science and Technology Major Project, China (2021SHZDZX0102), and National Science Foundation of China (Grant No. 61902247).

\clearpage
%
%
\bibliographystyle{splncs04}
\bibliography{ref}

\begin{thebibliography}{10}
\providecommand{\url}[1]{\texttt{#1}}
\providecommand{\urlprefix}{URL }
\providecommand{\doi}[1]{https://doi.org/#1}

\bibitem{azadi2020compositional}
Azadi, S., Pathak, D., Ebrahimi, S., Darrell, T.: Compositional gan: Learning
  image-conditional binary composition. International Journal of Computer
  Vision  \textbf{128},  2570--2585 (2020)

\bibitem{chen2019toward}
Chen, B.C., Kae, A.: Toward realistic image compositing with adversarial
  learning. In: CVPR (2019)

\bibitem{chen2009sketch2photo}
Chen, T., Cheng, M.M., Tan, P., Shamir, A., Hu, S.M.: Sketch2photo: Internet
  image montage. ACM transactions on graphics (TOG)  \textbf{28},  1--10 (2009)

\bibitem{cong2021bargainnet}
Cong, W., Niu, L., Zhang, J., Liang, J., Zhang, L.: Bargainnet:
  Background-guided domain translation for image harmonization. In: ICME (2021)

\bibitem{cong2022high}
Cong, W., Tao, X., Niu, L., Liang, J., Gao, X., Sun, Q., Zhang, L.:
  High-resolution image harmonization via collaborative dual transformations.
  In: CVPR (2022)

\bibitem{cong2020dovenet}
Cong, W., Zhang, J., Niu, L., Liu, L., Ling, Z., Li, W., Zhang, L.: Dovenet:
  Deep image harmonization via domain verification. In: CVPR (2020)

\bibitem{dai2019transformerxl}
Dai, Z., Yang, Z., Yang, Y., Carbonell, J., Le, Q.V., Salakhutdinov, R.:
  Transformer-xl: Attentive language models beyond a fixed-length context
  (2019)

\bibitem{deng2009imagenet}
Deng, J., Dong, W., Socher, R., Li, L.J., Li, K., Fei-Fei, L.: Imagenet: A
  large-scale hierarchical image database. In: CVPR (2009)

\bibitem{gehring2017convolutional}
Gehring, J., Auli, M., Grangier, D., Yarats, D., Dauphin, Y.N.: Convolutional
  sequence to sequence learning. In: ICML (2017)

\bibitem{georgakis2017synthesizing}
Georgakis, G., Mousavian, A., Berg, A.C., Kosecka, J.: Synthesizing training
  data for object detection in indoor scenes (2017)

\bibitem{goodfellow2014generative}
Goodfellow, I., Pouget-Abadie, J., Mirza, M., Xu, B., Warde-Farley, D., Ozair,
  S., Courville, A., Bengio, Y.: Generative adversarial nets. NIPS  (2014)

\bibitem{heusel2017gans}
Heusel, M., Ramsauer, H., Unterthiner, T., Nessler, B., Hochreiter, S.: Gans
  trained by a two time-scale update rule converge to a local nash equilibrium.
  NeurIPS  (2017)

\bibitem{hong2022shadow}
Hong, Y., Niu, L., Zhang, J.: Shadow generation for composite image in
  real-world scenes. In: AAAI (2022)

\bibitem{johnson2018image}
Johnson, J., Gupta, A., Fei-Fei, L.: Image generation from scene graphs. In:
  CVPR (2018)

\bibitem{kingma2014autoencoding}
Kingma, D.P., Welling, M.: Auto-encoding variational bayes (2014)

\bibitem{lalonde2007using}
Lalonde, J.F., Efros, A.A.: Using color compatibility for assessing image
  realism. In: ICCV (2007)

\bibitem{lalonde2007photo}
Lalonde, J.F., Hoiem, D., Efros, A.A., Rother, C., Winn, J., Criminisi, A.:
  Photo clip art. ACM transactions on graphics (TOG)  \textbf{26},  3--es
  (2007)

\bibitem{lee2018contextaware}
Lee, D., Liu, S., Gu, J., Liu, M.Y., Yang, M.H., Kautz, J.: Context-aware
  synthesis and placement of object instances (2018)

\bibitem{li2019putting}
Li, X., Liu, S., Kim, K., Wang, X., Yang, M.H., Kautz, J.: Putting humans in a
  scene: Learning affordance in 3d indoor environments. In: CVPR (2019)

\bibitem{lin2018st}
Lin, C.H., Yumer, E., Wang, O., Shechtman, E., Lucey, S.: St-gan: Spatial
  transformer generative adversarial networks for image compositing. In: CVPR
  (2018)

\bibitem{liu2020arshadowgan}
Liu, D., Long, C., Zhang, H., Yu, H., Dong, X., Xiao, C.: Arshadowgan: Shadow
  generative adversarial network for augmented reality in single light scenes.
  In: CVPR (2020)

\bibitem{liu2021opa}
Liu, L., Zhang, B., Li, J., Niu, L., Liu, Q., Zhang, L.: {OPA}: Object
  placement assessment dataset. arXiv preprint arXiv:2107.01889  (2021)

\bibitem{liu2020learning}
Liu, X., Yu, H.F., Dhillon, I., Hsieh, C.J.: Learning to encode position for
  transformer with continuous dynamical model. In: ICML (2020)

\bibitem{niu2021making}
Niu, L., Cong, W., Liu, L., Hong, Y., Zhang, B., Liang, J., Zhang, L.: Making
  images real again: A comprehensive survey on deep image composition. arXiv
  preprint arXiv:2106.14490  (2021)

\bibitem{raffel2020exploring}
Raffel, C., Shazeer, N., Roberts, A., Lee, K., Narang, S., Matena, M., Zhou,
  Y., Li, W., Liu, P.J.: Exploring the limits of transfer learning with a
  unified text-to-text transformer (2020)

\bibitem{schuster2010perceiving}
Schuster, M.J., Okerman, J., Nguyen, H., Rehg, J.M., Kemp, C.C.: Perceiving
  clutter and surfaces for object placement in indoor environments. In: ICHR
  (2010)

\bibitem{shaw2018selfattention}
Shaw, P., Uszkoreit, J., Vaswani, A.: Self-attention with relative position
  representations (2018)

\bibitem{simonyan2015deep}
Simonyan, K., Zisserman, A.: Very deep convolutional networks for large-scale
  image recognition (2015)

\bibitem{smith1996blue}
Smith, A.R., Blinn, J.F.: Blue screen matting. In: SIGGRAPH (1996)

\bibitem{tan2018and}
Tan, F., Bernier, C., Cohen, B., Ordonez, V., Barnes, C.: Where and who?
  automatic semantic-aware person composition. In: WACV (2018)

\bibitem{tripathi2019learning}
Tripathi, S., Chandra, S., Agrawal, A., Tyagi, A., Rehg, J.M., Chari, V.:
  Learning to generate synthetic data via compositing. In: CVPR (2019)

\bibitem{tsai2017deep}
Tsai, Y.H., Shen, X., Lin, Z., Sunkavalli, K., Lu, X., Yang, M.H.: Deep image
  harmonization. In: CVPR (2017)

\bibitem{vaswani2017attention}
Vaswani, A., Shazeer, N., Parmar, N., Uszkoreit, J., Jones, L., Gomez, A.N.,
  Kaiser, {\L}., Polosukhin, I.: Attention is all you need. In: NeurIPS (2017)

\bibitem{wang2018high}
Wang, T.C., Liu, M.Y., Zhu, J.Y., Tao, A., Kautz, J., Catanzaro, B.:
  High-resolution image synthesis and semantic manipulation with conditional
  gans. In: CVPR (2018)

\bibitem{weng2020misc}
Weng, S., Li, W., Li, D., Jin, H., Shi, B.: Misc: Multi-condition injection and
  spatially-adaptive compositing for conditional person image synthesis. In:
  CVPR (2020)

\bibitem{wu2019gp}
Wu, H., Zheng, S., Zhang, J., Huang, K.: Gp-gan: Towards realistic
  high-resolution image blending. In: ACM Multimedia (2019)

\bibitem{xue2012understanding}
Xue, S., Agarwala, A., Dorsey, J., Rushmeier, H.: Understanding and improving
  the realism of image composites. ACM Transactions on graphics (TOG)
  \textbf{31},  1--10 (2012)

\bibitem{zhang2020learning}
Zhang, L., Wen, T., Min, J., Wang, J., Han, D., Shi, J.: Learning object
  placement by inpainting for compositional data augmentation. In: ECCV (2020)

\bibitem{zhang2020deep}
Zhang, L., Wen, T., Shi, J.: Deep image blending. In: WACV (2020)

\bibitem{zhang2018unreasonable}
Zhang, R., Isola, P., Efros, A.A., Shechtman, E., Wang, O.: The unreasonable
  effectiveness of deep features as a perceptual metric. In: CVPR (2018)

\bibitem{zhang2020and}
Zhang, S.H., Zhou, Z.P., Liu, B., Dong, X., Hall, P.: What and where: A
  context-based recommendation system for object insertion. Computational
  Visual Media  \textbf{6},  79--93 (2020)

\bibitem{zhu2015learning}
Zhu, J.Y., Krahenbuhl, P., Shechtman, E., Efros, A.A.: Learning a
  discriminative model for the perception of realism in composite images. In:
  ICCV. pp. 3943--3951 (2015)

\bibitem{zhu2017multimodal}
Zhu, J.Y., Zhang, R., Pathak, D., Darrell, T., Efros, A.A., Wang, O.,
  Shechtman, E.: Multimodal image-to-image translation by enforcing bi-cycle
  consistency. In: NeurIPS (2017)

\end{thebibliography}


\begin{thebibliography}{1}
\providecommand{\url}[1]{\texttt{#1}}
\providecommand{\urlprefix}{URL }
\providecommand{\doi}[1]{https://doi.org/#1}

\bibitem{heusel2017gans}
Heusel, M., Ramsauer, H., Unterthiner, T., Nessler, B., Hochreiter, S.: Gans
  trained by a two time-scale update rule converge to a local nash equilibrium.
  NeurIPS  (2017)

\bibitem{jaderberg2015spatial}
Jaderberg, M., Simonyan, K., Zisserman, A., et~al.: Spatial transformer
  networks. NIPS  (2015)

\bibitem{liu2021opa}
Liu, L., Zhang, B., Li, J., Niu, L., Liu, Q., Zhang, L.: {OPA}: Object
  placement assessment dataset. arXiv preprint arXiv:2107.01889  (2021)

\bibitem{tripathi2019learning}
Tripathi, S., Chandra, S., Agrawal, A., Tyagi, A., Rehg, J.M., Chari, V.:
  Learning to generate synthetic data via compositing. In: CVPR (2019)

\bibitem{zhang2020learning}
Zhang, L., Wen, T., Min, J., Wang, J., Han, D., Shi, J.: Learning object
  placement by inpainting for compositional data augmentation. In: ECCV (2020)

\bibitem{zhang2018unreasonable}
Zhang, R., Isola, P., Efros, A.A., Shechtman, E., Wang, O.: The unreasonable
  effectiveness of deep features as a perceptual metric. In: CVPR (2018)

\end{thebibliography}
\end{document}


\pagestyle{headings}
\mainmatter

\title{Supplementary for Learning Object Placement via Dual-path Graph Completion} 

\titlerunning{Learning Object Placement}
%
\author{Siyuan Zhou \and
Liu Liu \and
Li Niu* \and Liqing Zhang}
\authorrunning{S. Zhou et al.}
%
\institute{MoE Key Lab of Artificial Intelligence, Shanghai Jiao Tong University, China
\email{\{ssluvble,Shirlley,ustcnewly\}@sjtu.edu.cn}, \email{zhang-lq@cs.sjtu.edu.cn}}

\maketitle

\let\thefootnote\relax\footnotetext{*Corresponding Author}

\noindent In this supplementary file, we first introduce more implementation details of our GracoNet in Section~\ref{sec:details}. Then, we conduct extensive ablation studies in Section~\ref{sec:ab}. We also provide additional visualizations to verify our method in Section~\ref{sec:vis}. Finally, we discuss the limitations in Section~\ref{sec:limit}.

\section{Implementation Details}\label{sec:details}

\subsection{Image Pre-processing}\label{subsec:imgpro}

All images are resized to $256\times 256$ and normalized before they are fed into the network, \emph{i.e.}, $H=256$ and $W=256$. Note that we maintain the relative aspect ratio between foreground and background before and after they are resized. For example, we use $(\mathrm{fg}^w$, $\mathrm{fg}^h)$ and $(\mathrm{bg}^w$, $\mathrm{bg}^h)$ to represent the original sizes of foreground and background, respectively. If $\mathrm{fg}^w/\mathrm{fg}^h>\mathrm{bg}^w/\mathrm{bg}^h$, we first resize foreground to $(256,(256\cdot\mathrm{fg}^h\cdot\mathrm{bg}^w)/(\mathrm{fg}^w\cdot\mathrm{bg}^h))$, and then zero-pads it to $(256, 256)$ on the top side and bottom side evenly. If $\mathrm{fg}^w/\mathrm{fg}^h\leqslant\mathrm{bg}^w/\mathrm{bg}^h$, we first resize foreground to $((256\cdot\mathrm{fg}^w\cdot\mathrm{bg}^h)/(\mathrm{fg}^h\cdot\mathrm{bg}^w), 256)$, and then zero-pads it to $(256, 256)$ on the left side and right side evenly. Meanwhile, background images and annotated composite images are directly resized to $(256, 256)$.

\subsection{Transformation Function $\mathcal{F}_{\mathbf{t}}$ with Parameter $\mathbf{t}$}\label{subsec:trans}

In our problem, we consider transformation parameters $\mathbf{t}=[t^r,t^x,t^y]\in\mathcal{R}^{3}$ with three degrees of freedom. We first define $t^r\in(0,1)$ to represent the scaling ratio for the foreground object. After scaling, the height and the width of the new foreground region are $h=t^r H$ and $w=t^r W$. Since we do not change the aspect ratio of foreground after transformation, $w$ and $h$ are not independent. The scaled foreground object will be placed at a reasonable location $(x,y)$ over the background, where $(x,y)$ represents the background coordinate for the left top pixel point of the foreground region. We then define $t^x=\frac{x}{W-w}\in(0,1)$ and $t^y=\frac{y}{H-h}\in(0,1)$ to indicate the relative vertical and horizontal locations that the foreground object should be placed over the background scene. For the accessibility of back propagation for both $\mathbf{t}_u$ and $\mathbf{t}_s$ in our dual-path framework, we follow Spatial Transformer Network (STN)~\cite{jaderberg2015spatial} to apply an affine transformation $\mathcal{A}$ with parameter $\Theta$ to transform the foreground region. With a simple derivation, the parameter $\Theta$ in our problem should be a function of $\mathbf{t}$: \begin{eqnarray}\label{eqn:affine}\Theta(\mathbf{t})=\begin{pmatrix} 1/t^r & 0 & (1-2t^x)(1/t^r-1) \\ 0 & 1/t^r & (1-2t^y)(1/t^r-1) \end{pmatrix}.\end{eqnarray} By applying affine transformation to foreground image $\mathrm{I}^{\mathrm{fg}}$ and foreground mask $\mathrm{M}^{\mathrm{fg}}$, we obtain a transformed foreground image $\hat{\mathrm{I}}^{\mathrm{fg}}= \mathcal{A}(\mathrm{I}^{\mathrm{fg}}; \Theta)$ with a new mask $\mathrm{M}^{\mathrm{c}} = \mathcal{A}(\mathrm{M}^{\mathrm{fg}}; \Theta)$. The predicted composite image $\mathrm{I}^{\mathrm{c}}$ is then calculated by $\mathrm{I}^{\mathrm{c}}=\mathrm{M}^{\mathrm{c}}*\hat{\mathrm{I}}^{\mathrm{fg}}+(1-\mathrm{M}^{\mathrm{c}})*\mathrm{I}^{\mathrm{bg}}$, in which $*$ means element-wise product. Since $\Theta$ is a function of $\mathbf{t}$, we could also describe $\mathrm{I}^{\mathrm{c}}$ and $\mathrm{M}^{\mathrm{c}}$ in function forms $f^I$ and $f^M$ conditioned on $\mathbf{t}$: \begin{eqnarray}\label{eqn:comp}\begin{aligned}\mathrm{I}^{\mathrm{c}}&= f^I(\mathrm{I}^{\mathrm{bg}}, \mathrm{I}^{\mathrm{fg}}, \mathrm{M}^{\mathrm{fg}};\mathbf{t}),\\\mathrm{M}^{\mathrm{c}}&= f^M(\mathrm{M}^{\mathrm{fg}};\mathbf{t}).\end{aligned}\end{eqnarray} Finally, our transformation function $\mathcal{F}_{\mathbf{t}}$ is defined by \begin{eqnarray}\label{eqn:trans_define}\mathcal{F}_{\mathbf{t}}(\mathrm{I}^{\mathrm{bg}}, \mathrm{I}^{\mathrm{fg}}, \mathrm{M}^{\mathrm{fg}})\triangleq(\,f^I(\mathrm{I}^{\mathrm{bg}}, \mathrm{I}^{\mathrm{fg}}, \mathrm{M}^{\mathrm{fg}};\mathbf{t}),\,f^M(\mathrm{M}^{\mathrm{fg}};\mathbf{t})\,).\end{eqnarray}

As discussed in Section 3.2 in the main paper, given a labeled positive composite image $\mathrm{I}^{\mathrm{c}}_{\mathrm{pos}}$ with object mask $\mathrm{M}^{\mathrm{c}}_{\mathrm{pos}}$, we should calculate its ground-truth transformation parameters $\mathbf{t}_{\mathrm{gt}} = [t^r_{\mathrm{gt}}, t^x_{\mathrm{gt}}, t^y_{\mathrm{gt}}]$ for calculating reconstruction loss $\mathcal{L}^{rec}_{s}$. The procedure of obtaining $\mathbf{t}_{\mathrm{gt}}$ from annotation derives from the definition of transformation parameters, as described in the following. We first obtain the bounding box of the foreground region on $\mathrm{I}^{\mathrm{c}}_{\mathrm{pos}}$/$\mathrm{M}^{\mathrm{c}}_{\mathrm{pos}}$, denoted by $(x_{\mathrm{gt}}, y_{\mathrm{gt}}, w_{\mathrm{gt}}, h_{\mathrm{gt}})$. Then, the ground-truth transformation parameters $\mathbf{t}_{\mathrm{gt}}$ are calculated by $t^r_{\mathrm{gt}}=\max(\frac{w_{\mathrm{gt}}}{W},\frac{h_{\mathrm{gt}}}{H})$, $t^x_{\mathrm{gt}}=\frac{x_{\mathrm{gt}}}{W-w_{\mathrm{gt}}}$, and $t^y_{\mathrm{gt}}=\frac{y_{\mathrm{gt}}}{H-h_{\mathrm{gt}}}$. 

\subsection{Reconstruction Loss $\mathcal{L}^{rec}_s$}\label{subsec:rec}

Since we expect the supervised path to reconstruct $(\mathrm{I}^{\mathrm{c}}_{s},\mathrm{M}^{\mathrm{c}}_{s})$, the reconstruction loss $\mathcal{L}^{rec}_s$ is designed to force $\mathbf{t}_s$ to be close to the ground-truth $\mathbf{t}_{\mathrm{gt}}$. We define $\mathcal{L}^{rec}_s$ as a weighted mean squared error (\emph{i.e.}, Weighted MSE) between $\mathbf{t}_s$ and $\mathbf{t}_{\mathrm{gt}}$:\begin{eqnarray}\label{eqn:loss_rec}\mathcal{L}^{rec}_s=\frac{\alpha^r(t^r_s-t^r_{\mathrm{gt}})^2+ \alpha^x(t^x_s-t^x_{\mathrm{gt}})^2 + \alpha^y(t^y_s-t^y_{\mathrm{gt}})^2}{3}\end{eqnarray} with weight ${\bm\alpha}=[\alpha^r,\alpha^x,\alpha^y]$. Specifically, we adopt dynamic weight in our implementation, where ${\bm\alpha}$ can be described as a set of functions determined by variable $t^r_s$, that is, $\alpha^r=f^r(t^r_s)$, $\alpha^x=f^x(t^r_s)$, and $\alpha^y=f^y(t^r_s)$. $f^r(t^r_s)$ should be a monotonically increasing function and $f^x(t^r_s),f^y(t^r_s)$ should be monotonically decreasing functions when $t^r_s\in(0,1)$. The reason for this design is intuitive. When $t^r_s$ is small and close to $0$, we pay more attention to where the foreground object should be placed, instead of the scale of the foreground region. Conversely, when $t^r_s$ is large and close to $1$, the relative vertical and horizontal location $t^x_s$ and $t^y_s$ become less important and now the scale of foreground object becomes our main concern. Concretely, we define ${\bm\alpha}$ in the form of trigonometric functions:
$\alpha^r=\sin(\frac{\pi}{2}t^r_s)$, $\alpha^x=\cos(\frac{\pi}{2}t^r_s)$, and $\alpha^y=\cos(\frac{\pi}{2}t^r_s)$. We have also explored other variants of reconstruction loss and compared them with our choice in Section~\ref{subsec:ab_rec}.

\subsection{Evaluation Metrics}\label{subsec:metric}

As introduced in Section 4.1 in the main paper, we adopt user study, accuracy, and FID~\cite{heusel2017gans} to evaluate generation plausibility, and adopt LPIPS~\cite{zhang2018unreasonable} to evaluate generation diversity during inference. In the following, we will discuss about more details in these four metrics.

\paragraph{User Study.}\label{para:metric_us}

The user study is conducted with 20 voluntary participants. We compare the object placement generation results of TERSE, PlaceNet, and our proposed method. For a given pair of foreground and background during inference (\emph{i.e.}, a test sample), each method produces one composite image. Then, each participant  chooses the most reasonable one from these three composite images. Each method is then scored by the proportion of participants who choose it (\emph{w.r.t.} this test sample). Finally, we average the score among all test samples to obtain the user study score for each method.

\paragraph{Accuracy.}\label{para:metric_acc}

We extend SimOPA~\cite{liu2021opa} model to check the accuracy of object placement generation results. The extended model functions as a binary classifier that distinguishes between reasonable and unreasonable object placements. We define accuracy as the proportion of the generated composite images that are classified as positive by the binary classifier during inference. We have released the code and model of the binary classifier for evaluation, and we omit the details here. 

\paragraph{FID.}\label{para:metric_fid}

Fréchet Inception Distance (FID)~\cite{heusel2017gans} is a measure of similarity between two datasets of images. It was shown to correlate well with human judgement of visual quality and is most often used to evaluate the quality of samples of Generative Adversarial Networks. We calculate FID score between one set of composite images generated by the network and another set of ground-truth positive composite images in the OPA $\emph{test}$ set.

\paragraph{LPIPS.}\label{para:metric_lpips}

In Generative Adversarial Networks, LPIPS~\cite{zhang2018unreasonable} is commonly used to measure the perceptual similarity between two images. In this work, we adopt LPIPS to measure the generation diversity of models. For a given pair of foreground and background during inference (\emph{i.e.}, a test sample), we generate 10 different composite images by sampling the random vector for 10 times. We first compute LPIPS for all pairs of composite images among 10 generation results for each test sample, and then calculate the averaged LPIPS among all test samples. Since LPIPS reveals the difference between two images, a larger LPIPS score corresponds to a better generation diversity.

\begin{table}[t]
  \centering
  \caption{Ablation study on degree of annotation}
  \label{table:ab_spath}
  \begin{tabular}{p{32mm}<{\centering}|p{13mm}<{\centering}p{13mm}<{\centering}|p{20mm}<{\centering}}
	\hline
	\multirow{2}*{Annotation Degree} & \multicolumn{2}{c|}{\emph{Plausibility}} & \multicolumn{1}{c}{\emph{Diversity}} \\
	& acc.$\uparrow$ & FID$\downarrow$ & LPIPS$\uparrow$ \\
	\hline
	$\mathcal{P}_u$ & 0.637 & 68.17 & 0 \\
	$\mathcal{P}_u+\mathcal{L}^{cls}_s$ & 0.754 & 34.80 & 0.130 \\
	$\mathcal{P}_u+\mathcal{P}_s$ & 0.847 & 27.75 & 0.206 \\
	\hline
  \end{tabular}
\end{table}

\begin{table}[t]
  \centering
  \caption{Ablation study on using negative training samples}
  \label{table:ab_neg}
  \begin{tabular}{p{20mm}<{\centering}p{8mm}<{\centering}p{8mm}<{\centering}|p{12mm}<{\centering}p{12mm}<{\centering}|p{16mm}<{\centering}}
	\hline
	\multirow{2}*{Method} & \multirow{2}*{Pos} & \multirow{2}*{Neg} & \multicolumn{2}{c|}{\emph{Plausibility}} & \multicolumn{1}{c}{\emph{Diversity}} \\
	& & & acc.$\uparrow$ & FID$\downarrow$ & LPIPS$\uparrow$ \\
	\hline
	\multirow{2}*{TERSE~\cite{tripathi2019learning}} & $\checkmark$ & & 0.588 & 49.35 & 0 \\
	& $\checkmark$ & $\checkmark$ & 0.679 & 46.94 & 0 \\
	\hline
	\multirow{2}*{PlaceNet~\cite{zhang2020learning}} & $\checkmark$ & & 0.619 & 32.50 & 0.101 \\
	& $\checkmark$ & $\checkmark$ & 0.683 & 36.69 & 0.160 \\
	\hline
	\multirow{2}*{GracoNet} & $\checkmark$ & & 0.808 & 27.15 & 0.206 \\
	& $\checkmark$ & $\checkmark$ & 0.847 & 27.75 & 0.206 \\
	\hline
  \end{tabular}
\end{table}

\begin{table}[t]
  \centering
  \caption{Ablation study on different reconstruction losses}
  \label{table:ab_rec}
  \begin{tabular}{p{10mm}<{\centering}p{15mm}<{\centering}p{15mm}<{\centering}|p{12mm}<{\centering}p{12mm}<{\centering}|p{16mm}<{\centering}}
	\hline
	\multirow{2}*{Type} & \multirow{2}*{$\alpha^r$} & \multirow{2}*{$\alpha^x$ \& $\alpha^y$} & \multicolumn{2}{c|}{\emph{Plausibility}} & \multicolumn{1}{c}{\emph{Diversity}} \\
	& & & acc.$\uparrow$ & FID$\downarrow$ & LPIPS$\uparrow$ \\
	\hline
	L1 & 1 & 1 & 0.820 & 29.38 & 0.063 \\
	L2 & 1 & 1 & 0.833 & 29.27 & 0.069 \\
	L2 & $t^r_s$ & $1-t^r_s$ & 0.836 & 27.92 & 0.190 \\
	\hline
	L2 & $\sin(\frac{\pi}{2}t^r_s)$ & $\cos(\frac{\pi}{2}t^r_s)$ & 0.847 & 27.75 & 0.206 \\
	\hline
  \end{tabular}
\end{table}

\section{Ablation Studies}\label{sec:ab}

\subsection{Degree of Annotation.}\label{subsec:ab_anndegree}

Table~\ref{table:ab_spath} shows an ablation study on the degree of annotation we use. Without the supervised path, the model witnesses a sharp decrease in performance and falls into mode collapse. After we add the classification loss $\mathcal{L}^{cls}_s$ to assist with the discriminator, the model works better in plausibility and relieves mode collapse. Adopting the complete supervised path brings another performance jump in all metrics. These experiments prove that fully exploiting supervision is crucial in object placement learning. Our supervised path is just designed under this guidance. By constructing a bijection between the latent vector and the predicted composite image, the model could effectively overcome the mode collapse problem. The supervised path successfully guides the unsupervised path to generate more reasonable and diversified object placements.

\subsection{Using Negative Samples for Training}\label{subsec:ab_neg}

As introduced in Section 1 in the main paper, OPA dataset~\cite{liu2021opa} is the first object placement assessment dataset that contains composite images and their binary rationality labels indicating whether they are reasonable (positive sample) or not (negative sample) in terms of foreground object placement. As discussed in Section 4.2 in the main paper, baseline TERSE~\cite{tripathi2019learning} and PlaceNet~\cite{zhang2020learning} did not include negative samples in their method because they had been proposed before OPA dataset was released. For most experiments on OPA dataset, we fairly use positive samples and negative samples together for both baselines and our method (\emph{e.g.}, experiments in Section 4.2 in the main paper). In this section, we aim to investigate whether introducing negative samples is necessary or not. 

Table~\ref{table:ab_neg} shows an ablation study on whether to use negative training samples for different methods in the training stage. As illustrated, accuracy drops without the assistance of negative samples in all methods. It is because simultaneously using positive samples and negative samples balances the process of adversarial training and enables the discriminator to learn from a wider range of data distribution. By comparing different methods, we find that removing negative samples from training process has the largest impact on TERSE (about 0.09 accuracy drop) and the smallest impact on our method (about 0.04 accuracy drop). This proves the robustness of our method because only our method still performs reasonably with the absence of negative samples.

\subsection{Different Types of Reconstruction Losses}\label{subsec:ab_rec}

As discussed in Section~\ref{subsec:rec}, we use a weighted MSE with trigonometric dynamic weights as the reconstruction loss $\mathcal{L}^{rec}_s$ between $\mathbf{t}_s$ and $\mathbf{t}_{\mathrm{gt}}$. In Table~\ref{table:ab_rec}, we explore different types of reconstruction losses, including L1-loss, L2-loss, L2-loss with linear dynamic weights, and L2-loss with trigonometric dynamic weights (our method). L1-loss and L2-loss both perform badly in generation diversity, because the model can not pay more attention to learning location (\emph{resp.}, size) information when $t^r_s$ is small (\emph{resp.}, large). L2-loss with linear/trigonometric dynamic weights overcomes this weakness and dynamically changes its attention during training. Comparably, trigonometric weights work slightly better than linear weights, so the former type becomes our final choice.

\subsection{Advantages Against Baselines}\label{subsec:ab_advantage}

TERSE, PlaceNet, and our method all adopt an adversarial training strategy. The generator functions to produce transformation parameters that reasonably places the foreground object over the background scene to form a composite image. The discriminator works by distinguishing between reasonable composite images and unreasonable composite images. The most important difference between baselines and our method lies in two aspects: the generator design and the usage of positive composite images. 

\paragraph{Generator Design.}\label{para:adv_gen}

TERSE uses a shared backbone followed by two separate branches to encode heterogeneous features for foreground and background. Then these two kinds of features are concatenated and regressed to predict transformation parameters. PlaceNet uses two independent encoders to extract features for foreground and background, respectively. Foreground and background features are concatenated with different random vectors to predict various object placements via a shared decoder network. Meanwhile, a diversity loss is designed to preserve the pairwise distance between the predicted placements and the corresponding random vectors. Compared with generators in these baselines, our proposed generator contains a novel GCM module that treats the object placement task as a graph completion problem. The background is considered as different nodes with different locations/sizes, whereas the foreground is considered as unique node lacking for location/size. GCM aims to reasonably place the foreground node among different background nodes to complete the graph. As introduced in Section 3.1 in the main paper, GCM mainly consists of Node Extraction Head (NEH) and Placement Seeking Network (PSN). We have investigated the functionality of NEH and PSN via an ablation study in Table 4 in the main paper, which shows that both NEH and PSN are crucial in GCM. Besides, according to Table~\ref{table:ab_spath}, without using the supervised path, our simplified version $\mathcal{P}_u+\mathcal{L}^{cls}_s$ containing GCM can already beat TERSE and PlaceNet in generation plausibility, which proves the advantage of our GCM design.


\begin{figure*}[t]
  \centering
  \includegraphics[width=\linewidth]{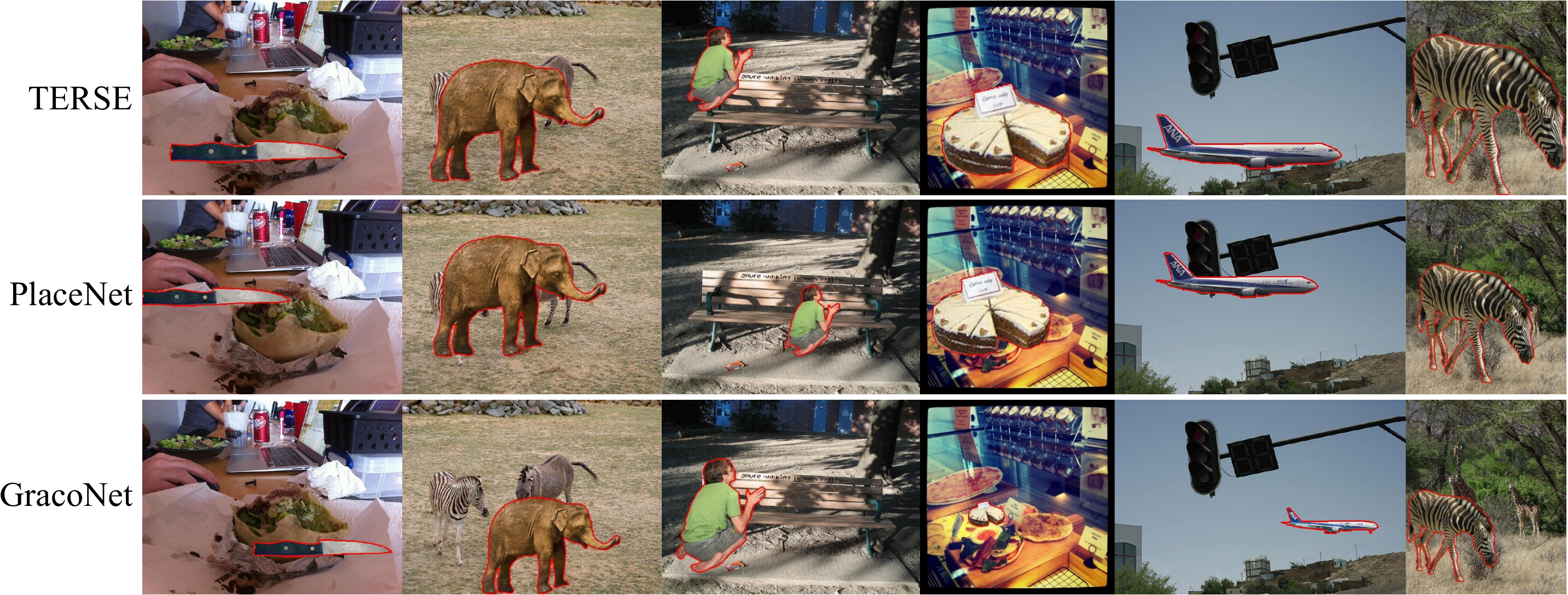}
  \caption{Visualization of object placement results for different foreground objects and background scenes on OPA \emph{test} set. Foreground is outlined in red}
  \label{fig:fgdiff_bgdiff_supp}
\end{figure*}

\begin{figure*}[t]
  \centering
  \includegraphics[width=\linewidth]{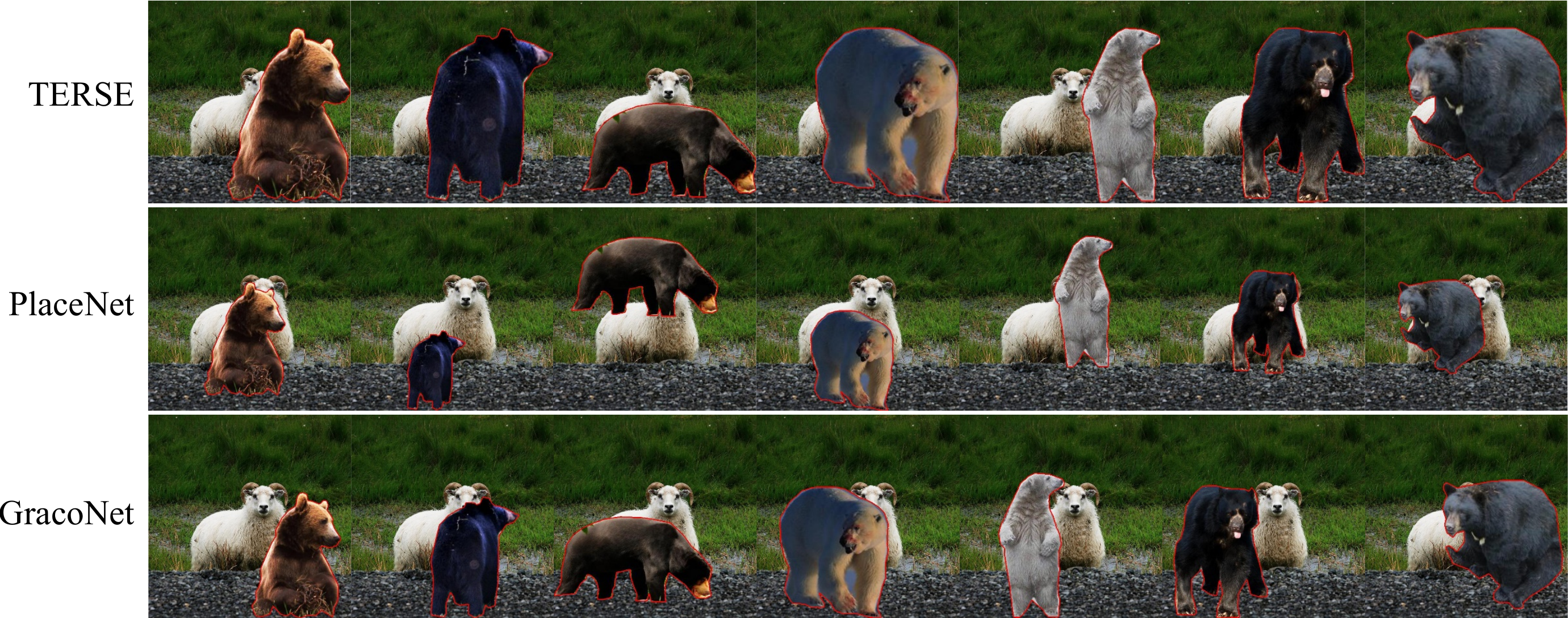}
  \caption{Visualization of object placement results for the same background scene with different foreground objects on OPA \emph{test} set. Foreground objects are outlined in red}
  \label{fig:fgdiff_bgsame}
\end{figure*}

\begin{figure*}[t]
  \centering
  \includegraphics[width=\linewidth]{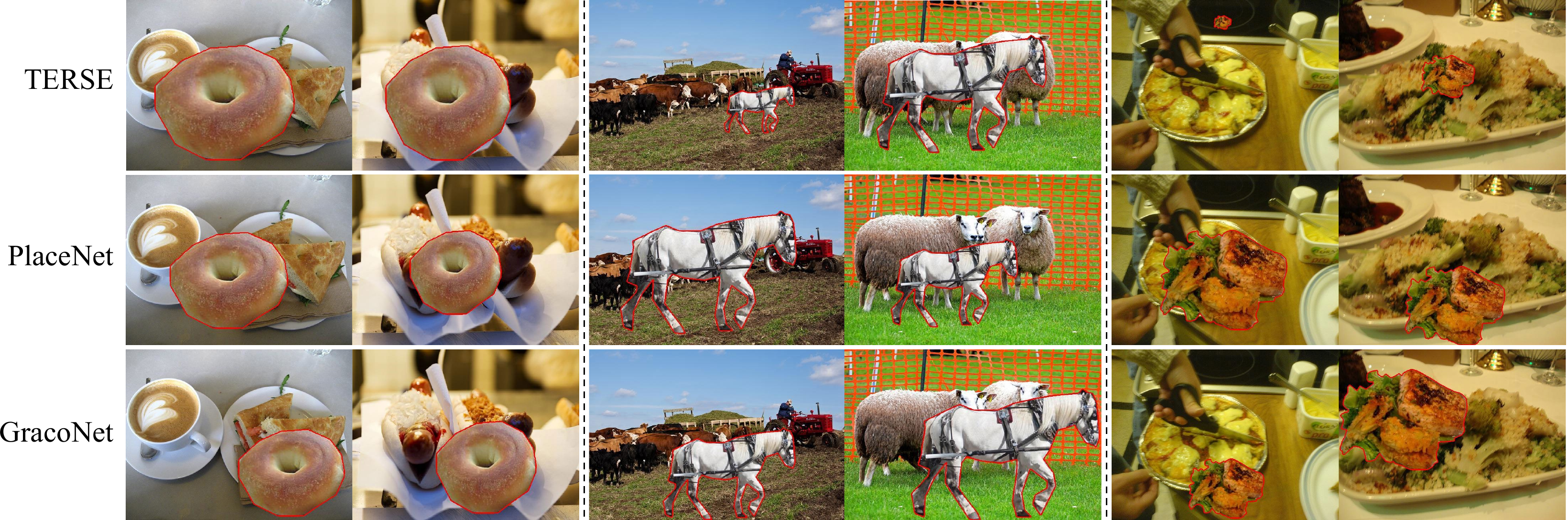}
  \caption{Visualization of object placement results for the same foreground object with different background scenes on OPA \emph{test} set. Foreground objects are outlined in red}
  \label{fig:fgsame_bgdiff}
\end{figure*}

\begin{figure*}[t]
  \centering
  \includegraphics[width=\linewidth]{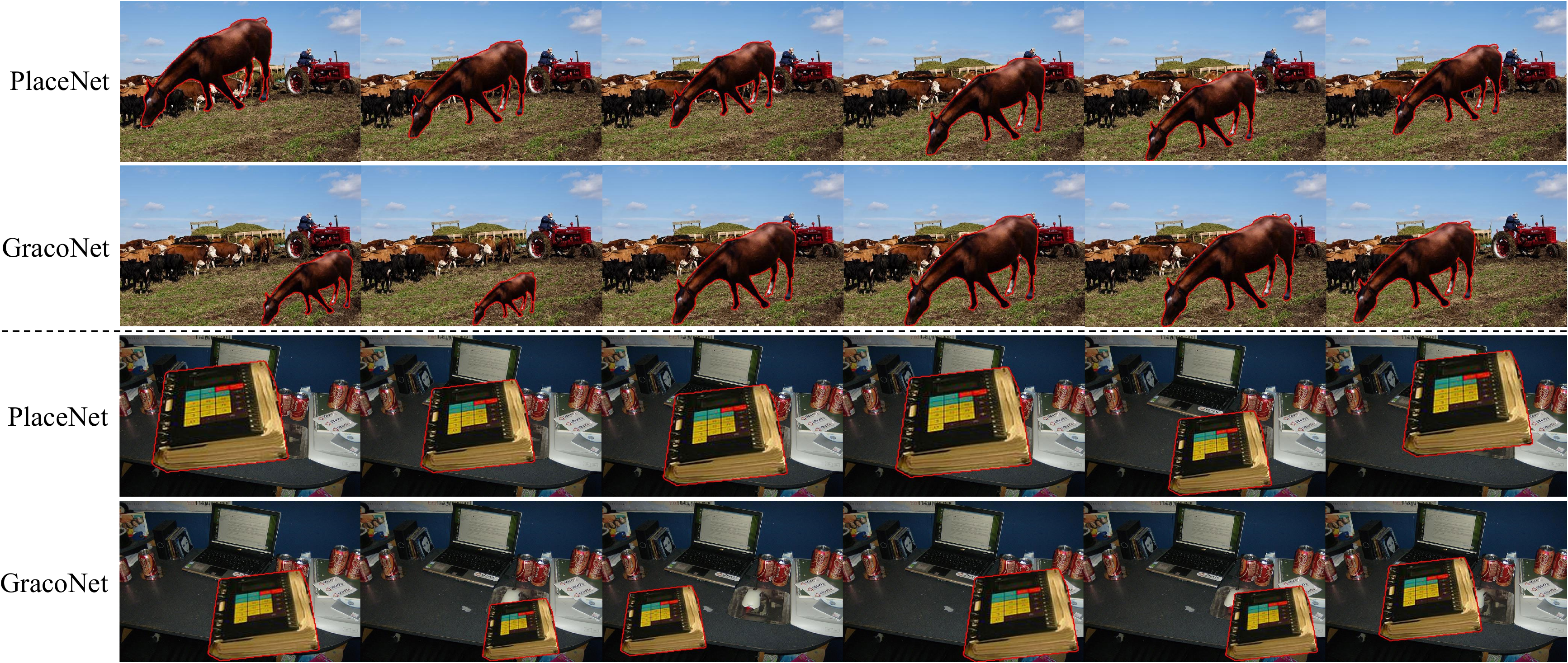}
  \caption{Visualization of object placement diversity on OPA \emph{test} set by sampling different random vectors. Foreground objects are outlined in red}
  \label{fig:fgsame_bgsame}
\end{figure*}

\paragraph{Usage of Positive Composite Images.}\label{adv_pos}

Baseline methods simply use positive composite images to train the discriminator, which wastes a lot of useful information. In contrast, we introduce a dual-path framework to effectively investigate the positive composite images. As introduced in Section 3.2 in the main paper, we establish a bijection between the latent vector and the predicted object placement in the supervised path. Specifically, we draw information from positive composite images to predict the latent vector, which is then utilized to reconstruct the positive composite image via a regression block and a transformation function.  Since the two paths share weights in most network layers, the supervised path could gradually guide the unsupervised path in the training stage. Under this design, our model successfully discovers the underlined object placement knowledge and produces satisfactory composite images during inference. By comparing $\mathcal{P}_u+\mathcal{L}^{cls}_s$ and $\mathcal{P}_u+\mathcal{P}_s$ in Table~\ref{table:ab_spath}, we can see that both generation plausibility and diversity are greatly enhanced after adding the supervised path, which demonstrates the power of supervised path in utilizing the positive composite images. 



\section{Visualization of Object Placement}\label{sec:vis}

Figure~\ref{fig:fgdiff_bgdiff_supp}, Figure~\ref{fig:fgdiff_bgsame}, Figure~\ref{fig:fgsame_bgdiff}, and Figure~\ref{fig:fgsame_bgsame} visualize more object placement results for different methods on OPA \emph{test} set, which supplement Figure 3 in the main paper. On the one hand, Figure~\ref{fig:fgdiff_bgdiff_supp}, Figure~\ref{fig:fgdiff_bgsame} and Figure~\ref{fig:fgsame_bgdiff} focus on generation plausibility of TERSE~\cite{tripathi2019learning}, PlaceNet~\cite{zhang2020learning}, and our method. Figure~\ref{fig:fgdiff_bgdiff_supp} displays the object placement results from different foreground objects and different background scenes. Figure~\ref{fig:fgdiff_bgsame} displays the combination of an identical background scene with different foreground objects. Figure~\ref{fig:fgsame_bgdiff} displays the combination of an identical foreground object with different background scenes. From these three figures, we find that our method not only adapts better to different foreground objects with reasonable locations and sizes conditioned on a given background, but also predicts more robust foreground objects under diverse background scenes.

On the other hand, Figure~\ref{fig:fgsame_bgsame} shows the generation diversity of PlaceNet~\cite{zhang2020learning} and our method by sampling different random vectors conditioned on the same pair of foreground and background. Our method outperforms PlaceNet by discovering more possible reasonable locations on the background, as well as changing spatial sizes of foreground correspondingly. In contrast, PlaceNet tends to meet mode collapse problems in some situations. These visualized examples effectively prove that our method simultaneously achieves better generation plausibility and diversity, and reaches a satisfactory balance between these two aspects.

\begin{figure*}[t]
  \centering
  \includegraphics[width=0.85\linewidth]{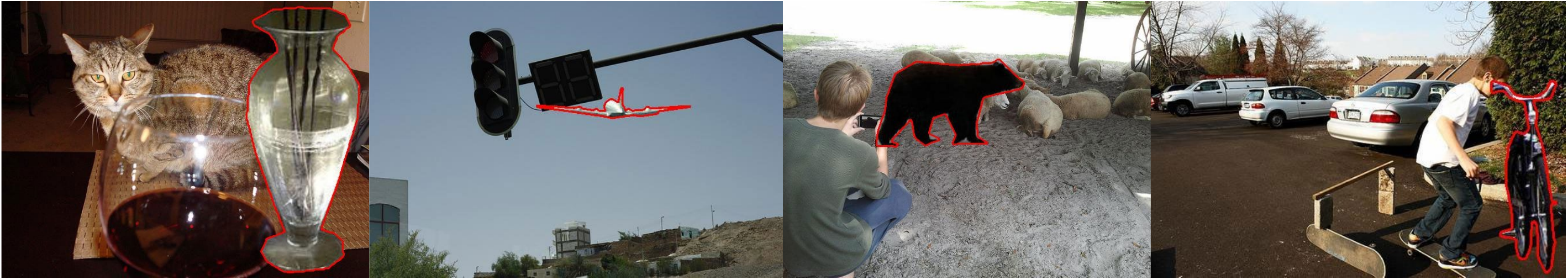}
  \caption{Failure cases in terms of occlusion between foreground and background}
  \label{fig:failure_case}
\end{figure*}

\section{Discussion on Limitation}\label{sec:limit}

In object placement, an important concern is occlusion between foreground and background. The generated composite images of a satisfactory object placement model should be reasonable when the foreground object is placed over the background scene. In TERSE and PlaceNet, we find that the foreground sometimes covers a counterintuitive background region, \emph{e.g.}, a sandwich wrongly covers the hand in row 2 and column 5 of Figure~\ref{fig:fgsame_bgdiff}. In our method, this phenomenon has been alleviated, but it still exists in some generated composite images, as shown in Figure~\ref{fig:failure_case}. This is probably because our model lacks a module that explicitly detects the occlusion relationship between foreground and background. GCM leverages the graph completion strategy to deal with the occlusion problem in an implicit way, which makes it surpass the baseline methods. However, incorporating a more explicit module to specifically address this problem may lead to better results. This provides a guiding direction for the optimization of future models on object placement.

\clearpage
%
%
\bibliographystyle{splncs04}
\bibliography{ref}